\documentclass{article} 
\usepackage{iclr2027_conference,times}

\usepackage{amsmath,amsfonts,bm}

\def\eqref#1{equation~\ref{#1}}

\def\1{\bm{1}}

\DeclareMathAlphabet{\mathsfit}{\encodingdefault}{\sfdefault}{m}{sl}
\SetMathAlphabet{\mathsfit}{bold}{\encodingdefault}{\sfdefault}{bx}{n}

\usepackage[utf8]{inputenc} 
\usepackage[T1]{fontenc}    
\usepackage{xcolor}         
\usepackage{graphicx}
\usepackage{url}            
\usepackage{booktabs}       
\usepackage{amsmath}
\usepackage{amsfonts}       
\usepackage{nicefrac}       
\usepackage{microtype}      
\usepackage{enumitem}
\usepackage{caption}
\usepackage{subcaption}
\usepackage{wrapfig}
\usepackage{dsfont}
\usepackage{adjustbox}
\usepackage{xspace}
\usepackage[para]{footmisc}
\usepackage[most]{tcolorbox}
\usepackage{multirow}
\usepackage{makecell}
\usepackage{tikz}
\usetikzlibrary{arrows.meta,positioning,calc,fit,backgrounds}
\usepackage{twemojis}
\usepackage[normalem]{ulem}

\definecolor{twblueback}{RGB}{247,247,252}
\definecolor{twblueframe}{RGB}{32,32,115}
\definecolor{tworangeback}{RGB}{255,250,245}
\definecolor{tworangeframe}{RGB}{166,106,0}
\definecolor{twgreenback}{RGB}{245,250,245}
\definecolor{twgreenframe}{RGB}{0,102,0}

\tcbset{
  tw/.style={
    enhanced, boxrule=0.5pt, arc=1.5mm,
    left=2mm, right=2mm, top=2.5mm, bottom=1mm,
    fonttitle=\bfseries\scriptsize, fontupper=\scriptsize,
    coltitle=white, attach boxed title to top left={xshift=2mm, yshift=-2mm},
    boxed title style={boxrule=0pt, arc=1mm},
  }
}

\usepackage[
    colorlinks=true,
    citecolor=blue,
    linkcolor=blue,
    urlcolor=blue
]{hyperref}                 

\definecolor{adibcolor}{RGB}{0,105,120}

\newcommand{\memo}{\mbox{\textsc{Memorilla}}\xspace}  
\newcommand{\gorilla}{\raisebox{-0.2ex}{\includegraphics[height=1em]{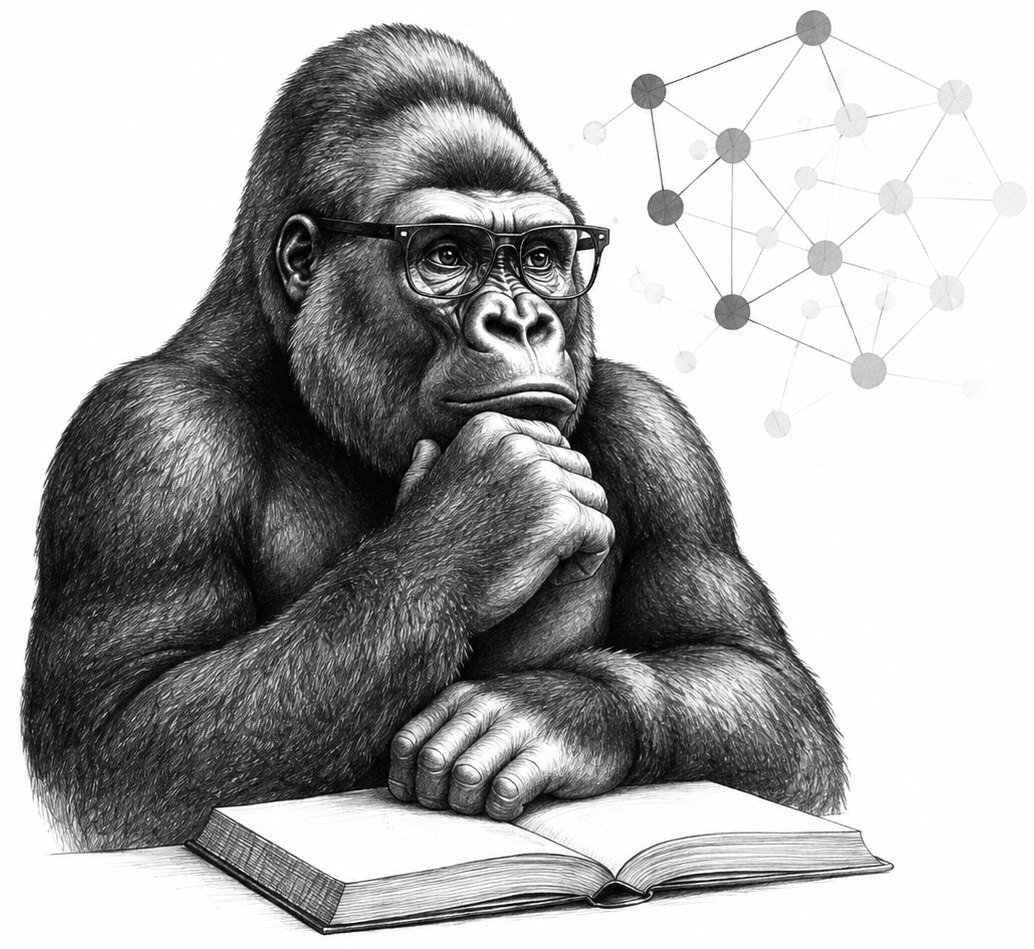}}}

\title{\gorilla~\memo:\\Latent Semantic Memory for LLMs}

\author{%
  \bfseries Adibvafa Fallahpour$^{1,2,5,*,\dagger}$ \quad Parsa Idehpour$^{3,*,\dagger}$ \quad Vignesh Kothapalli$^{1,*,\dagger}$ \\[3pt]
  \bfseries Nikita Mounier$^{3}$ \quad Shirley Wu$^{1}$ \quad Shayan Pardis$^{4}$ \quad Jure Leskovec$^{1,\dagger}$ \\[12pt]
  $^{1}$Stanford University \quad $^{2}$University of Toronto \quad $^{3}$University of Pennsylvania \\[1pt]
  $^{4}$MIT \quad $^{5}$Vector Institute \\[12pt]
  Website: \href{https://memory.stanford.edu/}{\texttt{https://memory.stanford.edu/}}
}

\iclrfinalcopy
\begin{document}

\maketitle
\lhead{Preprint}
\let\thefootnote\relax\footnotetext{
$^{*}$Equal contribution, sorted alphabetically.\\
$^{\dagger}$Correspondence to: 
\texttt{\{adib,vigneshk,jure\}}
  \texttt{@stanford.edu}, \texttt{ididea@seas.upenn.edu}
}

\begin{abstract}
Memory enables large language models (LLMs) to use information beyond their parameters, such as past user conversations, long documents, or an agent's earlier actions. For a frozen LLM, access to this accumulated knowledge is limited to what a memory system places in its context window. However, gathering and combining relevant information needed for each query while keeping the model's context short remains challenging for existing systems. Here we introduce \memo, a memory system that gathers information from an entire document collection into a fixed number of query-dependent `memory tokens' for a frozen LLM. \memo encodes documents (such as user chats, book chapters) once into reusable embeddings, then learns to retrieve and place query-specific information in the LLM's input embedding space. Using only 16 memory tokens, \memo achieves the best average performance against baselines including Mem0 and MemoRAG across benchmarks spanning personalization, long-document and knowledge-intensive question answering. Notably, on FactKG, it reaches 84.33\% accuracy versus 80.43\% with full-context inference, while using 108$\times$ fewer tokens to represent the source documents. Furthermore, \memo can facilitate continual learning in agentic systems where the document count (the agent's past turns) increases throughout the trajectory. In a TextWorld environment, we show that \memo reduces the agent's tendency to wander between rooms it has already explored, and improves its average success rate from 18.6\% with full-history in-context learning to 24.2\%. Overall, \memo aids frozen LLMs in accumulating and using knowledge effectively across long-context interactions and takes a step towards building continually learning AI systems.
\end{abstract}


\section{Introduction}
\label{sec:introduction}


Memory is central to how large language models (LLMs) are used. Personal assistants must recall a user's history across months of conversations, question-answering systems must reason over documents far longer than the model's context window, and agents must act on everything they have observed. None of this information exists in the model's parameters, and a frozen LLM can use it only through its context~\citep{hu2025memory}. A \emph{memory system} bridges this gap by encoding accumulated information, storing it persistently, and retrieving what each query needs into the context~\citep{wu2025human}. The central challenge for such a system is to give a frozen LLM access to an entire, growing collection of knowledge while keeping its context short~\citep{an2025why, du2025context}.

Existing memory systems address these issues only in part. Agentic memory systems such as MemGPT~\citep{packer2023memgpt}, Mem0~\citep{chhikara_khant_aryan_singh_yadav_2025} and A-MEM~\citep{xu2025mem} organize past interactions into summaries or structured records, and MemoRAG~\citep{qian2025memorag} uses a compressed global memory to guide retrieval, but each ultimately places selected text in the context and inherits the limits of selection. Latent compression methods such as ICAE~\citep{ge2024incontext}, IC-Former~\citep{wang2024context}, AutoCompressors~\citep{chevalier2023adapting} and ComMer~\citep{zeldes2025commer} represent documents as a few soft tokens in the LLM embedding space, but they form these tokens without the query, so the same tokens must serve every question. xRAG~\citep{cheng2024xrag} incorporates query relevance but only by compressing a single retrieved document. Methods that store model-specific hidden or key--value states, such as LongMem~\citep{wang2023augmenting}, MemoryLLM~\citep{wang2024memoryllm}, M+~\citep{wang2025m}, and Cartridges~\citep{eyuboglu2026cartridges}, tie the memory to one model and require training the LLM, a side network, or a separate cache for each corpus. What is missing is a memory system that will retrieve query-relevant information from document embeddings directly into the input embedding space of a frozen
LLM. Benefits of such an approach would be that the frozen LLM can retain all of its capabilities, and rely on only a few query relevant latent embeddings to answer questions or solve tasks instead of processing long textual contexts in a slow and lossy manner~\citep{liu2024lost}. In essence, improving the speed and quality of LLM responses and also the decision making in LLM agents based on past experiences. 

\begin{figure}[t!]
\centering
\includegraphics[width=0.9\linewidth]{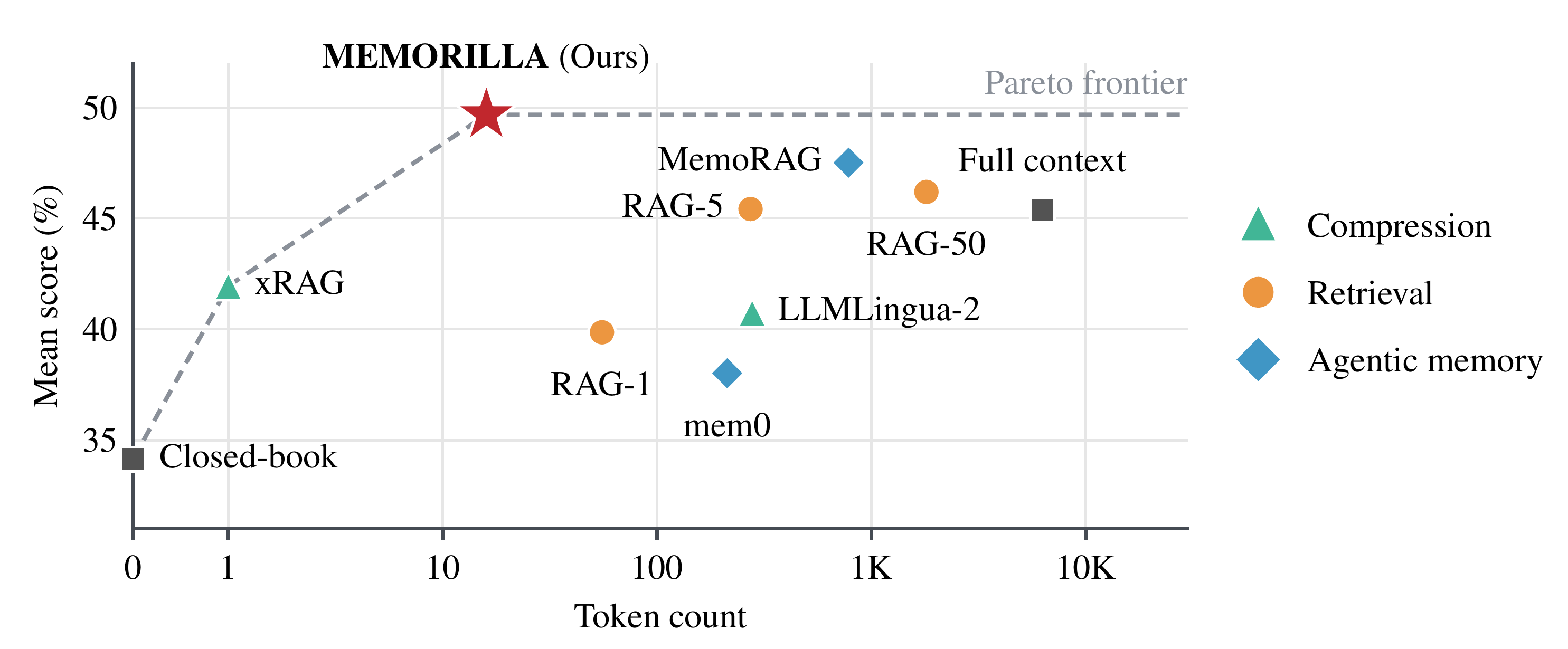}
\caption{\textbf{Performance versus context length.} Mean score over eight benchmarks against the geometric mean of document tokens given to a frozen Qwen3-8B decoder. \memo lies on the Pareto frontier (dashed) with the highest score at 16 memory tokens, 394$\times$ fewer than full context.
}
\label{fig:main_results}
\end{figure}

\textbf{Present work.} 
Here we introduce \memo, a memory system that learns to retrieve query-relevant information from document embeddings directly into the input embedding space of a frozen LLM (Figure~\ref{fig:architecture}). First, a frozen embedding model encodes each document, such as a user conversation or a book chapter, once into a persistent memory store. 
Next, for each query, a trainable memory module 
attends to every embedding in the persistent memory store and performs a query conditioned memory readout for the LLM. Unlike RAG which retrieves a subset of documents as textual tokens, \memo learns to recall what matters from the entire collection for a query, and expresses it as $K=16$ latent memory tokens for a frozen LLM, and does not require the documents to enter the context as text.
Because the store uses the same embeddings as a dense retrieval index, encoding adds no cost over retrieval, and with 512 documents on one H100, \memo has 27$\times$ lower end-to-end latency than full context answer generation (0.74s against 19.9s, Appendix~\ref{sec:compt}).

Our benchmarks mirror three settings in which deployed LLMs need memory (Figures~\ref{fig:main_results} and~\ref{fig:per-benchmark}). In personalization, an assistant must adapt to preferences a user rarely states directly, whether answering questions that depend on traits revealed across past conversations (PersonaMem-v2, PersonalizationV4) or writing in the style of the user's earlier text (LaMP-4, LaMP-7). Because a user's preferences rarely share wording with the query, similarity-based retrieval struggles to find and combine relevant information. Notably, on LaMP-4, \memo reaches 18.15 ROUGE-L against 14.21 for full context while using 474$\times$ fewer document tokens. In knowledge-intensive question answering, such as checking a claim against a knowledge base, \memo outperforms full context on FactKG (84.33\% against 80.43\%) with 108$\times$ fewer document tokens. In long-document understanding, the answer is buried in thousands of tokens of text and often phrased differently from the question, so a system must find and connect relevant content across the entire document. On TriviaQA, whose questions were written independently of their source pages, \memo reaches the highest exact match (57.00) with 743$\times$ fewer document tokens than full context.

Beyond answering questions, \memo allows agents to learn continually from their own experience. In this work, we consider the TextWorld environment~\citep{cote2018textworld} where an agent must explore rooms, find and use objects, and complete a multi-step goal within 50 turns, which requires remembering where it has been and what it has already tried. After every turn, \memo adds that turn to its document store and draws on this experience at the next turn. With our approach, the agent completes 37.5\% of held-out games in at least one of three attempts, against 25.0\% for the same agent given its full history in context. Overall, \memo improves the agent's ability to escape navigation loops and execute action sequences that are crucial to reaching the target.




\section{\memo}
\label{sec:method}

\begin{figure}[t]
\centering
\includegraphics[width=0.95\linewidth]{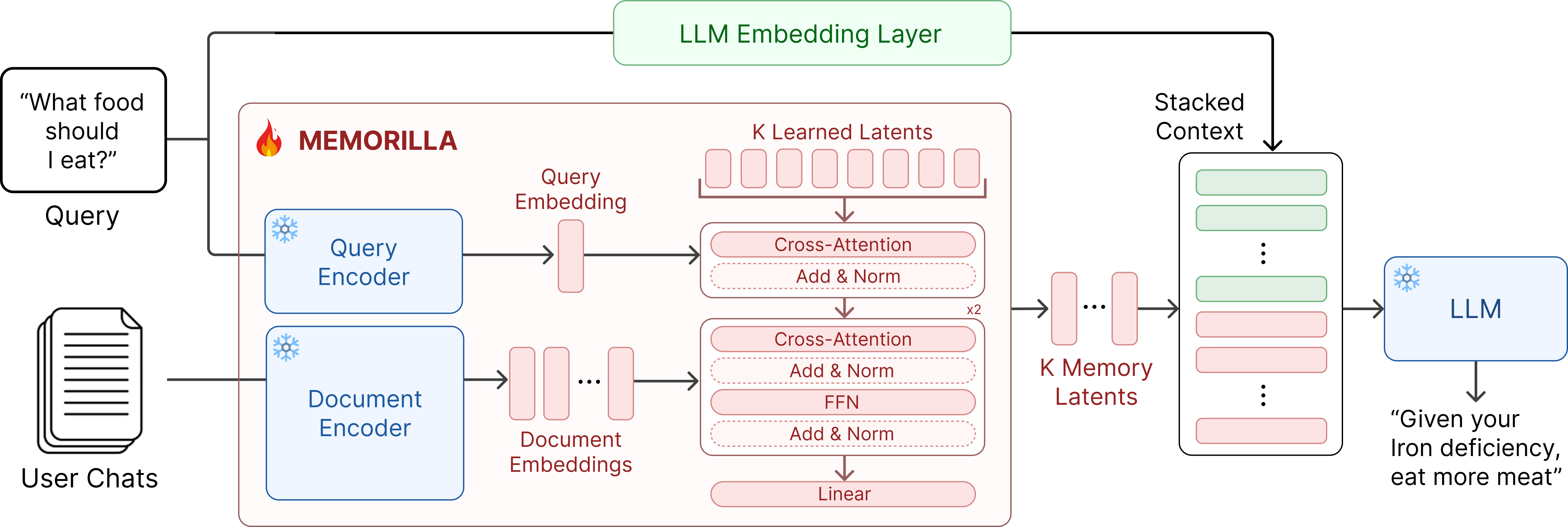}
\caption{\textbf{\memo architecture.} A frozen encoder embeds source documents (User Chats) once into a persistent memory store. For each Query, K query-conditioned slots attend over the store and are projected into the frozen LLM's embedding space as a memory readout.}
\label{fig:architecture}
\vspace{-3mm}
\end{figure}

A persistent latent memory must preserve a document collection across
queries while recalling different information for each query. \memo meets
these requirements through three operations: \emph{encoding},
\emph{storage}, and \emph{retrieval}. Given $N$ document chunks
$\mathcal{D}=\{d_1,\ldots,d_N\}$, a frozen encoder produces a
query-independent representation that is stored once. For instance, $\mathcal{D}$ can be a collection of $N$ user-LLM conversations, chapters of a lengthy book, or health records.
For each query $q$,
a trainable memory module reads this store and returns $K$ query-conditioned
soft memory tokens, which a frozen LLM consumes together with the query text. Figure~\ref{fig:architecture} summarizes the architecture.

\subsection{Encoding and Persistent Storage}

A frozen document encoder $f_{\mathrm{enc}}^d$ maps each chunk $d_i$ to an
embedding $\mathbf{z}_i\in\mathbb{R}^{h_e}$. Stacking these embeddings
produces the query-independent document store
\begin{align}
    \mathbf{Z}_{\mathcal D}
    = [f_{\mathrm{enc}}^d(d_1),\ldots,f_{\mathrm{enc}}^d(d_N)]
    \in \mathbb{R}^{N\times h_e}.
\end{align}
Because $\mathbf{Z}_{\mathcal D}$ does not depend on a query, it is computed
once and persisted in memory or on secondary storage. Every subsequent
query reuses the same matrix, and only added or changed documents must be
re-encoded. 
A frozen query encoder $f_{\mathrm{enc}}^q$ then maps $q$ to
$\mathbf{z}_q=f_{\mathrm{enc}}^q(q)\in\mathbb{R}^{h_e}$. 
Unlike $\mathbf{Z}_{\mathcal D}$, $\mathbf{z}_q$ is computed only for the current query and is not stored as long-term memory.

\subsection[Query-Conditioned Retrieval]{Query-Conditioned Retrieval
($f_{\text{mem}}$)}

In the next phase, a trainable memory module $f_{\text{mem}}$ uses $\mathbf{z}_q$ to select information from
$\mathbf{Z}_{\mathcal D}$:
\begin{align}
    f_{\text{mem}}:
    (\mathbf{Z}_{\mathcal D},\mathbf{z}_q)
    \mapsto \mathbf{M}, \qquad
    \mathbf{M}\in\mathbb{R}^{K\times h_L},
\end{align}
where $h_L$ is the input-embedding dimension of the frozen LLM and
$\mathbf{M}=[\mathbf{m}_1;\ldots;\mathbf{m}_K]$ is the output memory
matrix. Each row $\mathbf{m}_k\in\mathbb{R}^{h_L}$ is one soft memory token
in the LLM embedding space. The module reads all $N$ stored document
embeddings but always contributes exactly $K$ memory tokens to the LLM
context, and allows significant context compression (over $4000\times$, see Table~\ref{tab:main-results}).

\textbf{Stage 1: Memory slot initialization.}
The module initializes $K$ memory slots with learned vectors
$\mathbf{Q}_0\in\mathbb{R}^{K\times h_e}$ that are shared across
collections and queries. Distinct initial values define a fixed output
budget while allowing the slots to capture different information.

\textbf{Stage 2: Query conditioning.}
The slots attend to the query embedding:
\begin{align}
    \mathbf{Q}_c = \text{LayerNorm}\!\left(
    \mathbf{Q}_0 + \text{CrossAttn}
    (Q{=}\mathbf{Q}_0,K{=}\mathbf{z}_q,V{=}\mathbf{z}_q)\right).
\end{align}
Because the query is represented by one vector, it provides a shared update
to all slots; the residual connection preserves the differences among
$\mathbf{Q}_0$. The resulting $\mathbf{Q}_c$ is therefore query-conditioned
while retaining distinct slot identities.

\textbf{Stage 3: Document retrieval.}
The conditioned slots attend to every document embedding through
$L_{\mathrm{ref}}$ cross-attention blocks, following the latent-array design of
Perceiver~\citep{jaegle_gimeno_brock_zisserman_vinyals_carreira_2021}. Let
$\overline{\mathbf{Z}}_{\mathcal D}
=\text{LayerNorm}(\mathbf{Z}_{\mathcal D})$ and
$\mathbf{R}_0=\mathbf{Q}_c$. For block $\ell$,
\begin{align}
\begin{split}
    \widetilde{\mathbf{R}}_\ell
    &= \mathbf{R}_{\ell-1}
    + \text{CrossAttn}\!\left(
      Q{=}\text{LayerNorm}(\mathbf{R}_{\ell-1}),
      K{=}\overline{\mathbf{Z}}_{\mathcal D},
      V{=}\overline{\mathbf{Z}}_{\mathcal D}\right), \\
    \mathbf{R}_\ell
    &= \widetilde{\mathbf{R}}_\ell
    + \text{FFN}\!\left(\text{LayerNorm}
      (\widetilde{\mathbf{R}}_\ell)\right),
\end{split}
\end{align}
where FFN is a two-layer feed-forward network with GELU
activations~\citep{hendrycks2016gaussian}. Each slot can
attend to the full collection, while the query-conditioned slots determine
which information is retained. The refinement blocks therefore implement a
continuous, query-conditioned read over all stored documents rather than a
discrete top-$k$ selection.

\textbf{Stage 4: Projection to the LLM embedding space.}
A learned projection maps the refined slots from the encoder dimension
$h_e$ to the LLM input-embedding dimension $h_L$:
\begin{align}
    \mathbf{M}
    = \text{LayerNorm}(\mathbf{R}_{L_{\text{ref}}})
      \mathbf{W}_{\text{proj}}, \qquad
    \mathbf{W}_{\text{proj}}\in\mathbb{R}^{h_e\times h_L}.
\end{align}
The resulting rows $\mathbf{m}_1,\ldots,\mathbf{m}_K$ are the soft memory
tokens passed to the frozen LLM.

\subsection{Memory Injection into the Frozen LLM}

Finally, the LLM input-embedding layer maps the text query to $\mathbf{P}_q$, giving
\begin{align}
\label{eq:memory_inj}
    \mathbf{X}_{\text{LLM}}
    = [\underbrace{\mathbf{m}_1,\ldots,\mathbf{m}_K}_{
       \text{soft memory tokens}};\mathbf{P}_q].
\end{align}
Because $\mathbf{m}_k\in\mathbb{R}^{h_L}$ has the same dimension as a text
token embedding, the LLM processes memory and text in the same input
sequence. No LLM layer or attention mechanism is changed. Rotary position
embeddings are applied over the combined sequence in the standard
way~\citep{su_lu_pan_murtadha_wen_liu_2021}. The frozen LLM can therefore
attend to the recalled memory through its existing self-attention layers.

\subsection{Supervised Training Objective}
\label{sec:training_objective}
The frozen LLM generates
$\mathbf{Y}_{\text{out}}=(y_1,\ldots,y_L)$ autoregressively from
$\mathbf{X}_{\text{LLM}}$, and we minimize the negative log-likelihood of
the target answer tokens:
\begin{align}
    \mathcal{L}_{\mathrm{sup}}
    = -\sum_{i=1}^{L}
      \log P(y_i\mid \mathbf{y}_{<i},\mathbf{X}_{\text{LLM}};
      \boldsymbol{\theta}_{\text{mem}}),
\end{align}
where $\boldsymbol{\theta}_{\text{mem}}$ includes the memory module and
its projection to the LLM embedding space. The loss is applied only to answer positions, not to
the prompt or memory positions. 

\subsection{\memo with Qwen3}

Based on the formalization above, we instantiate \memo with $f_{\mathrm{enc}}^d$ and $f_{\mathrm{enc}}^q$ as a frozen Qwen3-Embedding-4B model. Unless stated otherwise, we fix $K=16$ throughout the paper and use a Qwen3-8B model as the frozen LLM. This gives us 194{,}123{,}776 trainable parameters.


\subsection{Training Stages}
\label{sec:training_datasets}
We use a three-stage curriculum that moves from general content preservation
to diverse language tasks and then to the evaluation domains. See Appendix~\ref{app:hyperparameters} for training hyperparameters.

\textbf{Stage 1} trains the memory module on 1.98 million English Wikipedia
paraphrase reconstruction examples. Each passage is divided into document
chunks, and the module must preserve enough information in its fixed memory
budget for the frozen LLM to generate a paraphrase. This stage provides a
general initialization before the module encounters task-specific
supervision.

\textbf{Stage 2} continues training on 1.34 million examples drawn from ten datasets
covering reading comprehension, open-domain question answering, and
summarization. The variation in source structure and answer format teaches
the module to recall different information from the same type of stored
representation. Table~\ref{tab:training-data} lists the component datasets
and their sizes.

\textbf{Stage 3} fine-tunes the resulting module on a unified mixture of the training
splits from all eight evaluation benchmarks. One multi-task checkpoint is
used directly for FactKG, TriviaQA, NarrativeQA, PubMedQA, LaMP-4, and
LaMP-7. PersonaMem-v2 and PersonalizationV4 receive additional
benchmark-specific fine-tuning because their validation losses required more epochs to plateau, while the other benchmarks plateaued or began to overfit earlier.




\section{Experiments}
\label{sec:experiments}
\subsection{Baselines}

We compare \memo against RAG approaches based on cosine similarity, Mem0~\citep{chhikara_khant_aryan_singh_yadav_2025}, and MemoRAG~\citep{qian2025memorag}. xRAG~\citep{cheng2024xrag} and ComMer~\citep{zeldes2025commer} (via its reported results, Table~\ref{tab:commer}) compress the context into latent input embeddings for the LLM. Finally, we also compare against LLMLingua-2~\citep{llmlingua2}, which is a task-agnostic text based prompt compression method. All baselines we run use a frozen Qwen3-Embedding-4B encoder for document embeddings and a frozen Qwen3-8B decoder for generation. Mem0 and MemoRAG additionally use a frozen Qwen3-4B-Instruct-2507 to build their memories, matching the 4B scale of the encoder with which \memo reads the same collections (Appendix~\ref{app:baselines}).

\subsection{Benchmarks}

We evaluate \memo on eight supervised benchmarks spanning (a) personalization, (b) long-document QA, and (c) knowledge-intensive QA tasks that demand memory. For personalization, we consider PersonaMem-v2 (PMv2) \citep{jiang_yuan_shen_hao_xu_chen_liu_vijjini_he_yu_et_al_2025}, LaMP-4, LaMP-7 \citep{salemi_mysore_bendersky_zamani_2023} and our own PersonalizationV4 (PV4) benchmarks as they require integrating information scattered across a user's prior chats or writing history to form a coherent user representation. For long-document QA, we choose NarrativeQA \citep{kovcisky2018narrativeqa} and TriviaQA~\citep{joshi_choi_weld_zettlemoyer_2017} to test the method's ability to compress long documents in their entirety into a usable representation for downstream QA. Finally, for knowledge-intensive QA, we consider FactKG \citep{kim_park_kwon_jo_thorne_choi_2023} and PubMedQA \citep{jin2019pubmedqadatasetbiomedicalresearch} to test factual grounding from structured or semi-structured evidence of varying quantity and format.
Table~\ref{tab:context-budget} (Appendix~\ref{app:context-budget}) reports how many document tokens each method gives the LLM per query across all the benchmarks.



\subsection{Evaluation}

We use accuracy for classification tasks, exact match and token-level F1 (lowercased, whitespace-tokenized) for short-form QA, and ROUGE, BLEU, F1, and METEOR for open-ended text generation. PMv2 and PV4 are evaluated on their short-answer questions. The model's generation is embedded with the frozen encoder and scored as correct if the gold answer is its cosine nearest neighbor among the question's predefined answer choices, which we report as short-answer accuracy. All supervised LLM evaluations use a decoding temperature of 0. Our serving stack enforces a $32{,}768$-token decoder context and enables YaRN whenever an assembled full-context input exceeds this limit (Appendix~\ref{app:benchmark_details}).


\section{Quantitative Results}
\label{sec:quantitative}

\begin{figure}[t]
\centering
\includegraphics{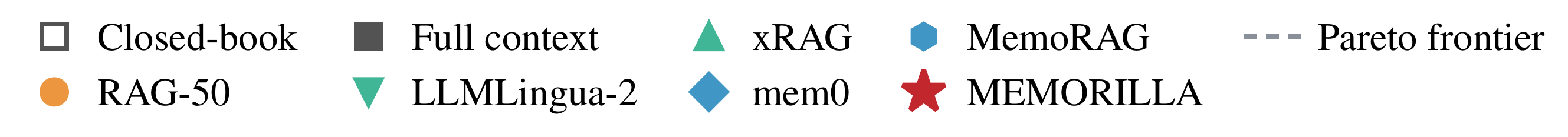}\par\smallskip
\begin{subfigure}[t]{1.36in}\centering
  \includegraphics{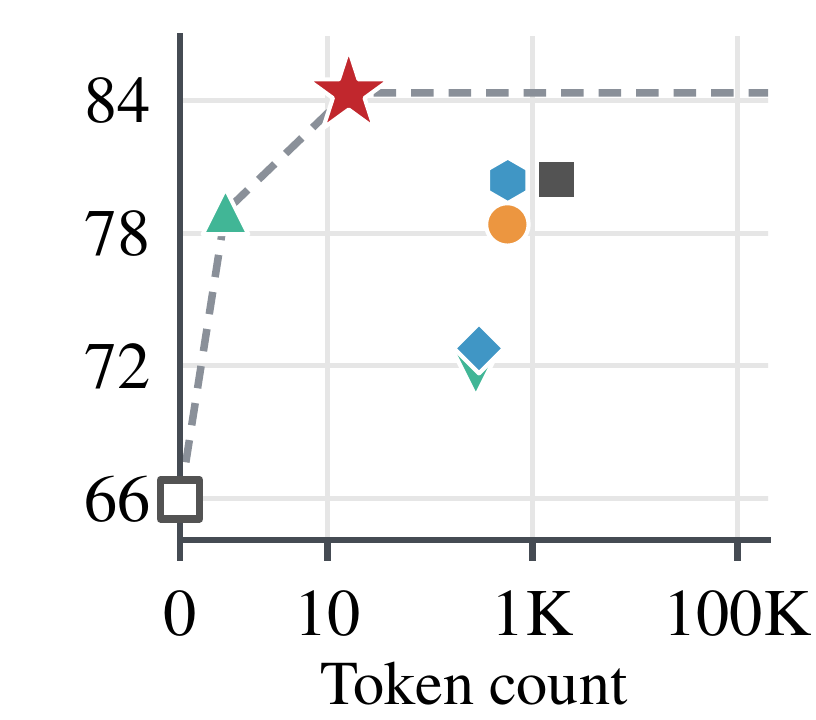}
  \caption{FactKG (Acc)}\label{fig:per-benchmark-factkg}
\end{subfigure}\hfill
\begin{subfigure}[t]{1.36in}\centering
  \includegraphics{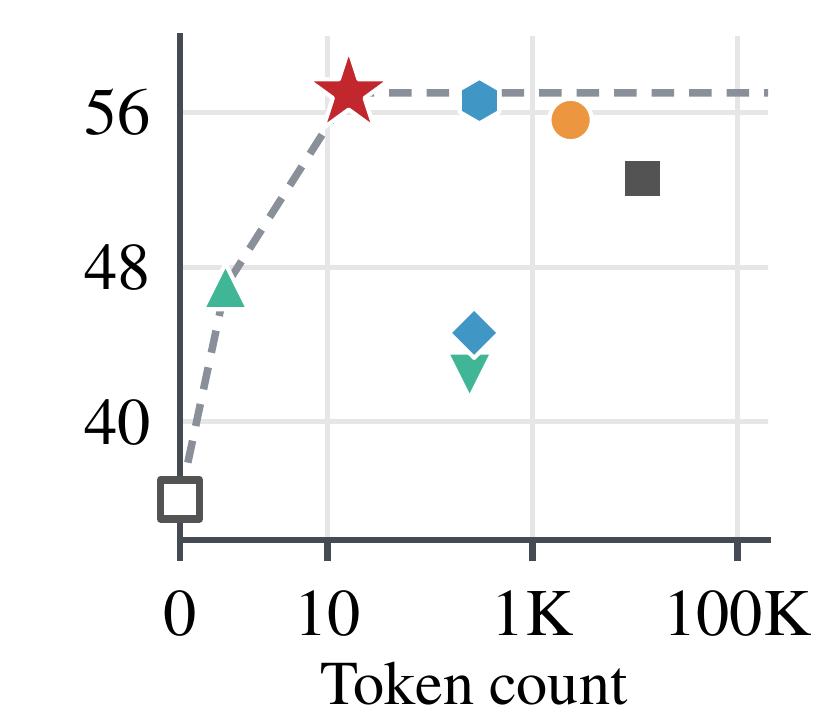}
  \caption{TriviaQA (EM)}\label{fig:per-benchmark-triviaqa}
\end{subfigure}\hfill
\begin{subfigure}[t]{1.36in}\centering
  \includegraphics{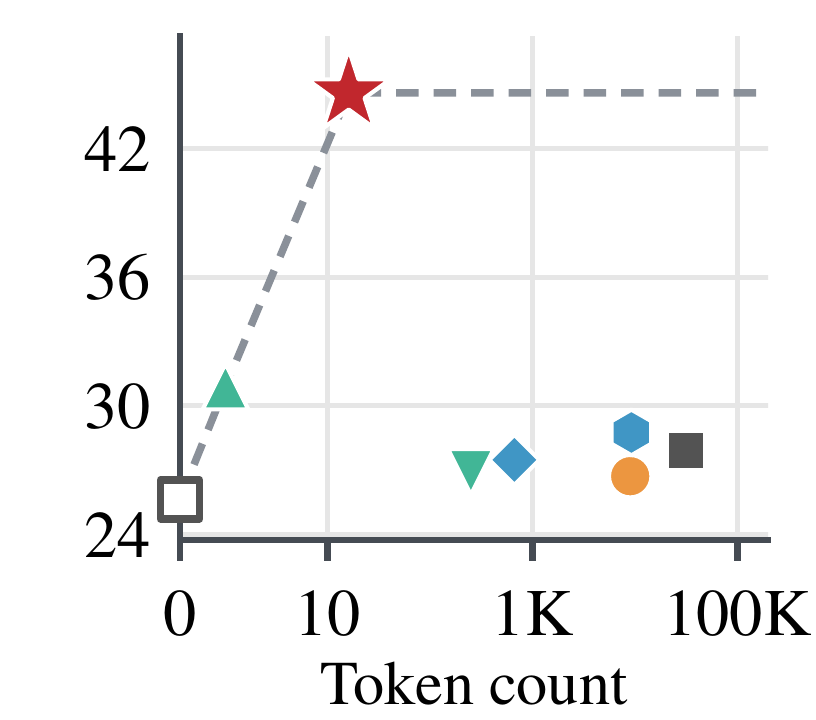}
  \caption{PMv2 (Acc)}\label{fig:per-benchmark-pmv2}
\end{subfigure}\hfill
\begin{subfigure}[t]{1.36in}\centering
  \includegraphics{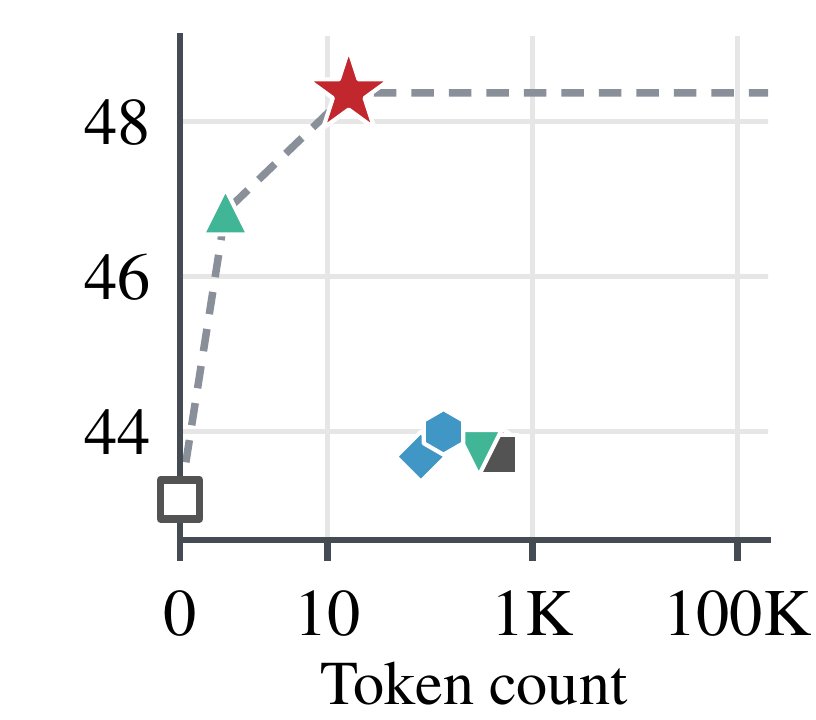}
  \caption{LaMP-7 (ROUGE-L)}\label{fig:per-benchmark-lamp7}
\end{subfigure}\par\medskip
\begin{subfigure}[t]{1.36in}\centering
  \includegraphics{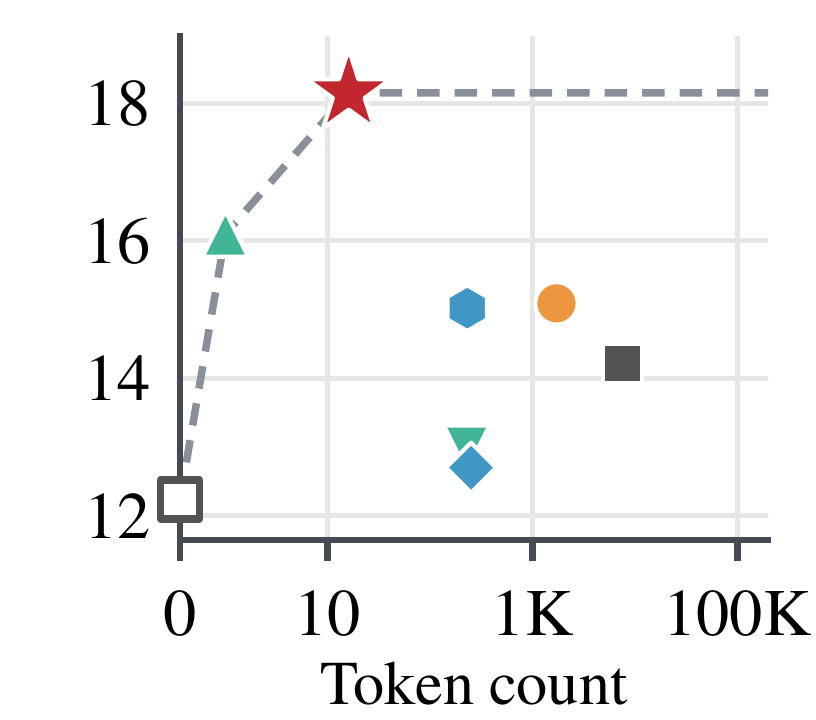}
  \caption{LaMP-4 (ROUGE-L)}\label{fig:per-benchmark-lamp4}
\end{subfigure}\hfill
\begin{subfigure}[t]{1.36in}\centering
  \includegraphics{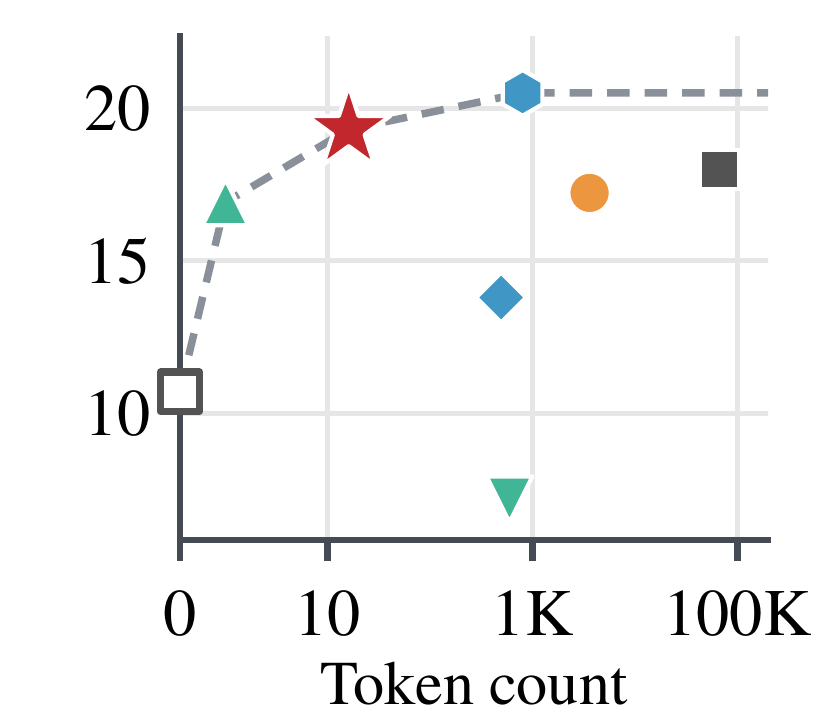}
  \caption{NQA (ROUGE-L)}\label{fig:per-benchmark-narrativeqa}
\end{subfigure}\hfill
\begin{subfigure}[t]{1.36in}\centering
  \includegraphics{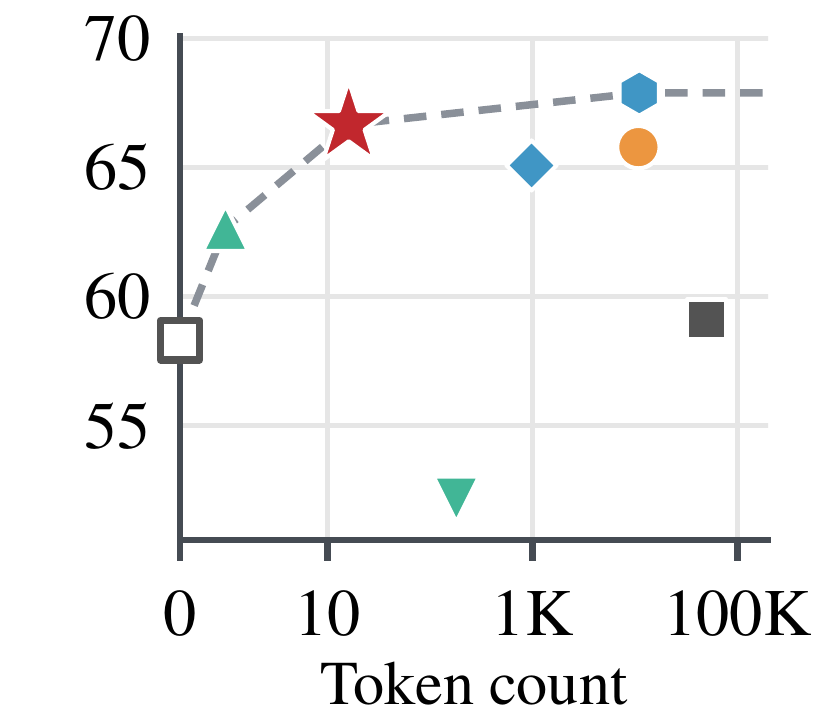}
  \caption{PV4 (Acc)}\label{fig:per-benchmark-pv4}
\end{subfigure}\hfill
\begin{subfigure}[t]{1.36in}\centering
  \includegraphics{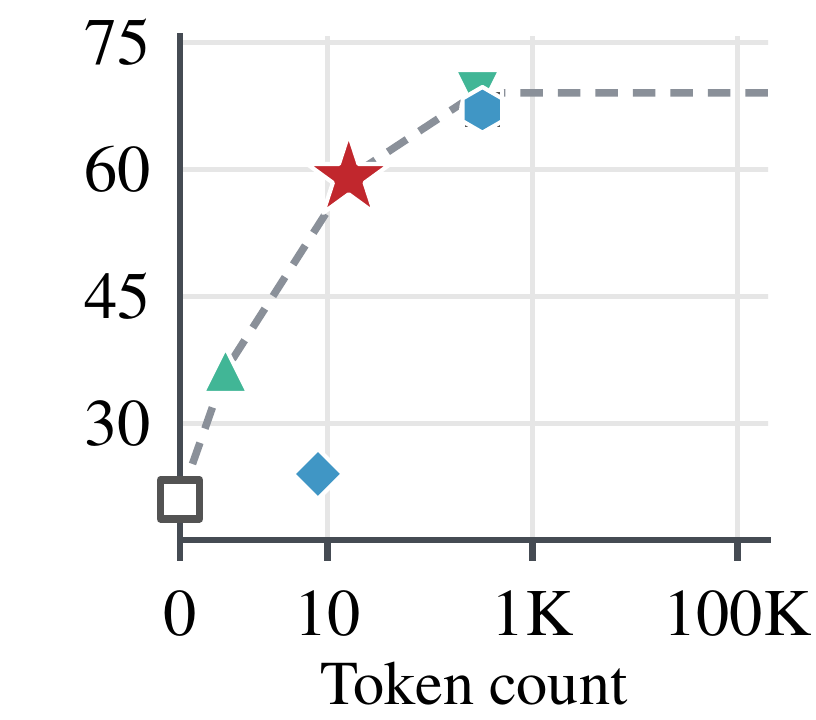}
  \caption{PubMedQA (Acc)}\label{fig:per-benchmark-pubmedqa}
\end{subfigure}
\caption{\textbf{Per-benchmark performance versus context length.} Primary-metric score against mean document tokens per query, with a frozen Qwen3-8B decoder. \memo lies on the Pareto frontier (dashed) of every benchmark, leads on five and is second on NarrativeQA and PV4.
}
\label{fig:per-benchmark}
\vspace{-4mm}
\end{figure}



\memo outperforms all baselines on average while using 394$\times$, 113$\times$ and 49$\times$ fewer document tokens than full context, RAG-50 and MemoRAG in geometric mean (Figure~\ref{fig:main_results}). It also lies on the Pareto frontier of every benchmark, while the compact baselines xRAG, LLMLingua-2 and Mem0 average only 41.92, 40.75 and 38.02 against 49.68. Figure~\ref{fig:per-benchmark} breaks this down by benchmark, where \memo leads on five and and is second on NarrativeQA and PV4.

\textbf{Relational evidence.} On FactKG (Figure~\ref{fig:per-benchmark-factkg}), \memo reaches 84.33\% against 80.43\% for full context, 80.38\% for MemoRAG and 78.98\% for xRAG, with 108$\times$ and 36$\times$ fewer tokens than the first two. Because full context receives every document, the margin cannot stem from discarded evidence, and Section~\ref{sec:what-learns} shows that \memo's margin over RAG-50 grows with the number of entities and widens overall with the number of composed facts.



\textbf{Long-document and personalization QA.} On TriviaQA (Figure~\ref{fig:per-benchmark-triviaqa}), \memo reaches 57.00 EM against 56.60 for MemoRAG and 52.60 for full context with 19$\times$ and 743$\times$ fewer document tokens. On PMv2 short-answer (Figure~\ref{fig:per-benchmark-pmv2}), \memo reaches 44.60\% against 30.90\% for xRAG, the strongest baseline. Every other baseline stays within about three points of closed-book (25.64\%), including full context, which uses 1{,}974$\times$ more document tokens than \memo. On LaMP-7, every baseline except xRAG (46.84) stays within one ROUGE-L point of the closed-book model, whereas \memo improves it from 43.13 to 48.37. Finally, \memo achieves 18.15 ROUGE-L on LaMP-4 (Figure~\ref{fig:per-benchmark-lamp4}) with 474$\times$ fewer tokens than the full context baseline.

\textbf{Where memory does not help.} On the remaining three benchmarks (Figures~\ref{fig:per-benchmark-narrativeqa}--\ref{fig:per-benchmark-pubmedqa}), a baseline leads, but only by supplying the LLM with 18 to 690 times more document tokens.  MemoRAG leads NarrativeQA in \mbox{ROUGE-1} and \mbox{ROUGE-L} (21.80 and 20.51 versus 19.68 and 19.31) with 50$\times$ more document tokens, consistent with its access to verbatim passages, and PV4 (67.88\% versus 66.65\%) with 690$\times$ more. On PubMedQA, whose four abstracts total only 324 tokens, LLMLingua-2 reaches 69.00\% and four other baselines 67.00\%, versus 59.00\% for \memo.


\textbf{Full context is not an upper bound.} Full context trails \memo on every benchmark except PubMedQA and on 20 of 21 metrics in Table~\ref{tab:main-results}, and RAG-50 exceeds it on 15, so supplying the entire collection hurts more often than it helps.

\section{What Does \memo Learn?}
\label{sec:what-learns}


\begin{figure}[!t]
\centering
\captionsetup[sub]{skip=2pt}  
\includegraphics{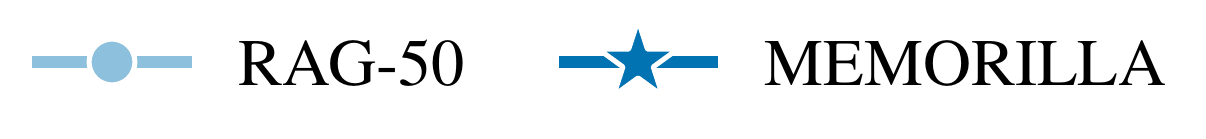}\par\vspace{-2mm}
\begin{subfigure}[t]{2.2in}\centering
  \includegraphics{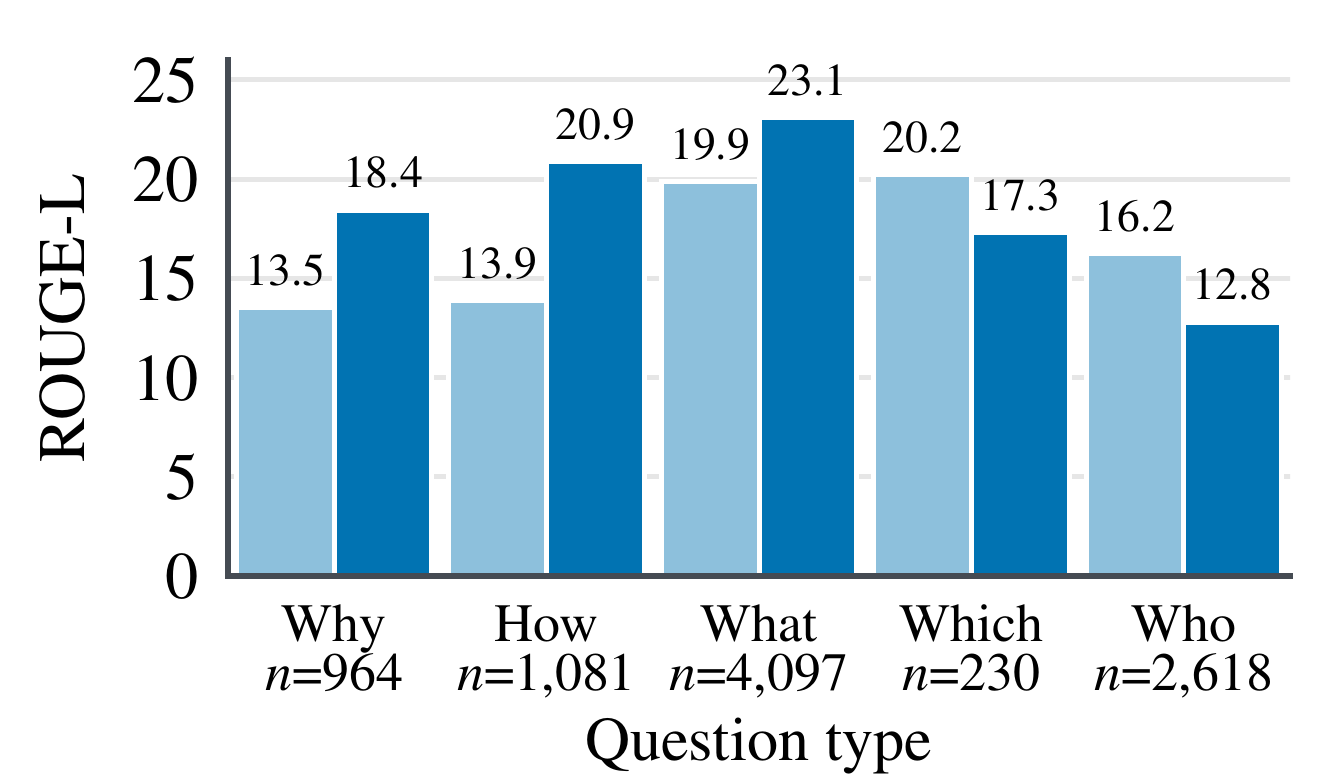}
  \caption{NarrativeQA by question type}\label{fig:nqa-qtype}
\end{subfigure}\hfill
\begin{subfigure}[t]{1.75in}\centering
  \includegraphics{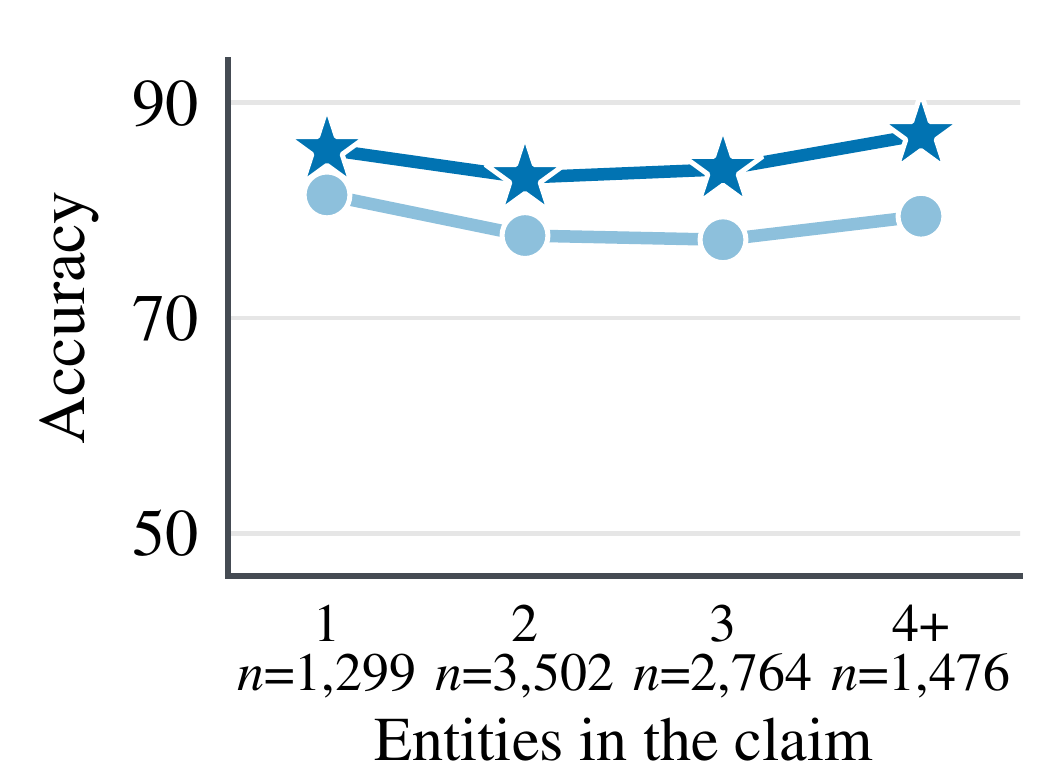}
  \caption{FactKG by entity count}\label{fig:factkg-entities}
\end{subfigure}\hfill
\begin{subfigure}[t]{1.55in}\centering
  \includegraphics{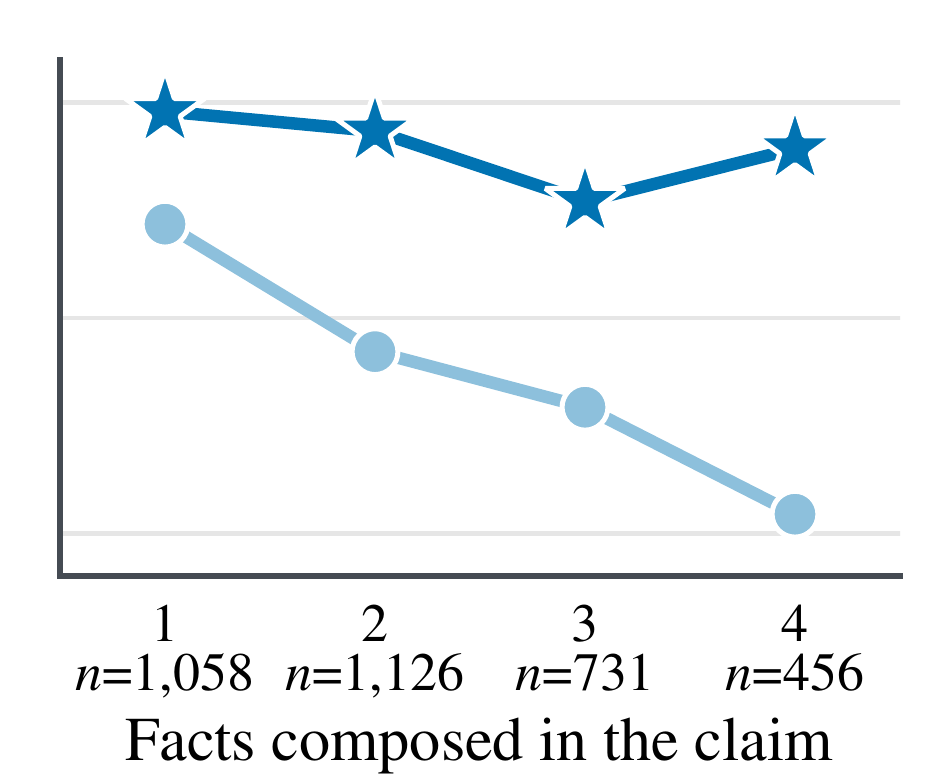}
  \caption{FactKG by composition}\label{fig:factkg-composition}
\end{subfigure}
\caption{\textbf{Per-question breakdowns against RAG-50.} \textbf{(a)} NarrativeQA question types are the question's first word. Conceptual questions (`Why', `How', `What') favor \memo whereas entity-recall  questions (`Which', `Who') favor RAG. \textbf{(b)} \memo's advantage grows with entity count, indicating better capture of multi-entity relational patterns. \textbf{(c)} RAG-50 degrades as more facts are composed, while \memo remains stable across all depths.}
\label{fig:breakdowns}
\vspace{-2mm}
\end{figure}

\begin{figure}[t!]
\vspace{10pt}
\centering
\includegraphics[width=0.9\textwidth]{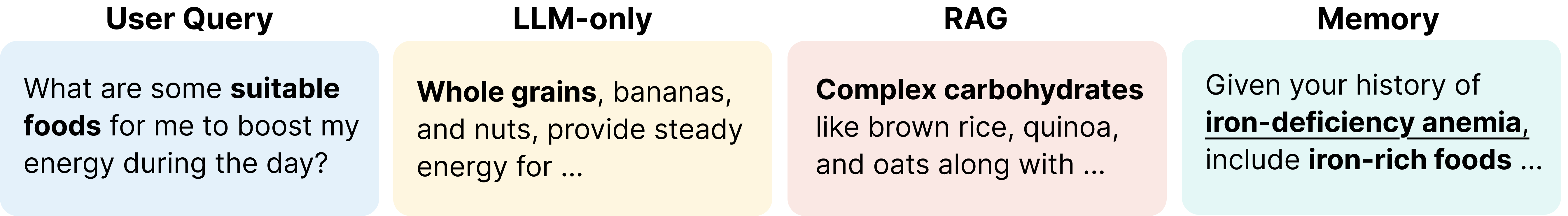}
\caption{\textbf{Personalization Case Study.} On PersonaMem-v2, \memo personalizes a generic query by leveraging the user's remembered iron-deficiency anemia history, despite no lexical overlap between the question and the relevant conversation history.}
\label{fig:case-study}
\end{figure}

\textbf{Conceptual understanding vs. factual recall.} Figure~\ref{fig:nqa-qtype} breaks down NarrativeQA performance by question type. \memo outperforms RAG-50 by +4.91 ROUGE-L on ``why'' questions and +7.02 on ``how,'' both requiring reasoning about narrative structure and causality. Conversely, RAG-50 leads by +3.45 on ``who'' and by +2.93 on ``which,'' where retrieval can locate the entity name verbatim. One explanation is that compression preserves global structure for conceptual reasoning but loses the exact surface tokens that retrieval excels at pinpointing (although question type also varies with answer length and entity frequency). The same trade-off would explain why MemoRAG, which reads selected passages, leads our ROUGE scores on this benchmark while trailing on F1. Figure~\ref{fig:case-study} illustrates the complementary strength in personalization, where \memo uses semantically relevant medical history despite no lexical overlap with the query.

\textbf{Multi-entity relational reasoning.} On FactKG, \memo's advantage over RAG-50 grows monotonically with entity count, from +4.23pp at 1 entity to +7.59pp at 4+ (Figure~\ref{fig:factkg-entities}). RAG-50 drops from 78.73 to 51.75 as the number of composed facts increases from 1 to 4, while \memo remains in the 80--89 range (Figure~\ref{fig:factkg-composition}). Compression encodes relational structure holistically, whereas retrieval degrades when the reasoning chain spans multiple documents. Because 73.5\% of FactKG claims have more than 50 chunks, part of this gap reflects evidence RAG-50 never retrieves, but full context, which reads every chunk, still trails \memo overall (80.43 against 84.33).

\begin{figure}[t!]
\centering
\includegraphics[width=0.9\textwidth]{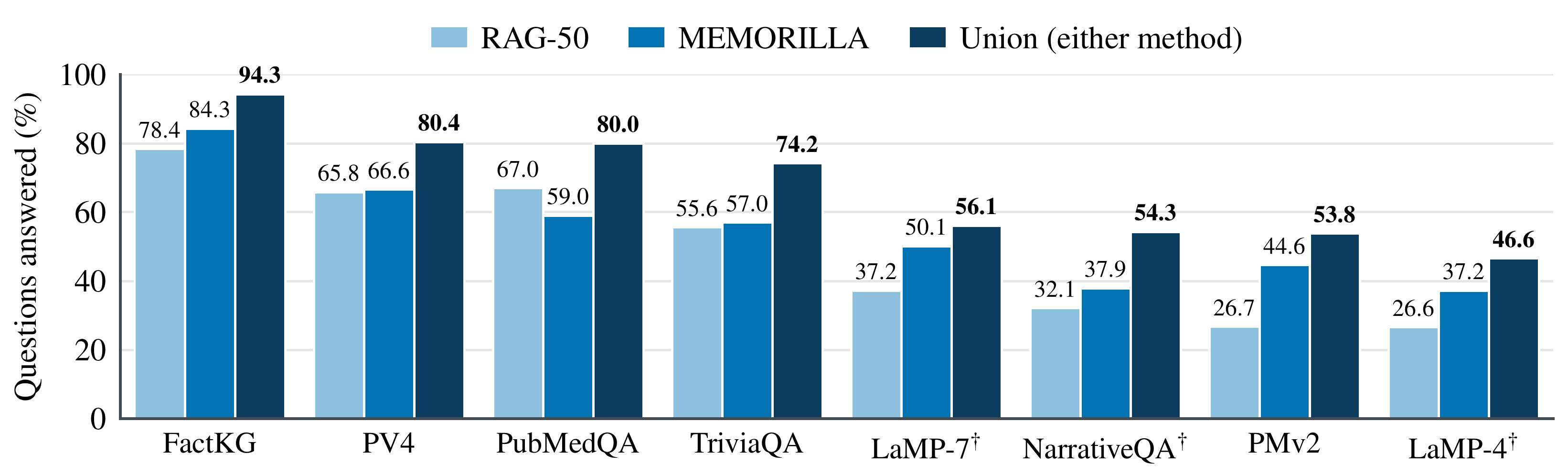}  
\caption{\textbf{Per-question complementarity between \memo{} and RAG-50.} Share of questions answered by each method alone and by at least one of them (union), an oracle upper bound for a hybrid system. The union exceeds either method on every benchmark. $^{\dagger}$A question counts as answered when its ROUGE-L is at least $\max(0.20, m)$, where $m$ is \memo's median.}
\label{fig:complementarity}
\vspace{-3mm}
\end{figure}

\textbf{\memo and RAG solve different questions.} On FactKG, we observed that 15.9\% of questions are answered by \memo alone against 10.0\% by RAG-50 alone, and on PMv2 short-answer the split is 27.1\% against 9.2\%. In Figure~\ref{fig:complementarity}, we show that the union of \memo and RAG (which is an oracle upper bound), 
could lead to substantial gains in performance than either approach alone, which motivates the design of hybrid retrieval and latent-memory systems as potential future work.

\textbf{What neither method solves.} We also examine the questions that both \memo and RAG-50 answer incorrectly. On PV4, many of these require linking facts that each appear somewhere in the user's history, which suggests that the remaining errors come from combining information rather than from failing to find it. On FactKG, both methods are biased toward judging claims false. True claims make up 89.5\% of the claims both methods get wrong but only 28.7\% of those both get right, which points to a miscalibrated decision threshold rather than missing evidence.

\section{Continual Learning with \memo}
\label{sec:rl}

\begin{figure}[t]
\centering
\includegraphics[width=1\linewidth]{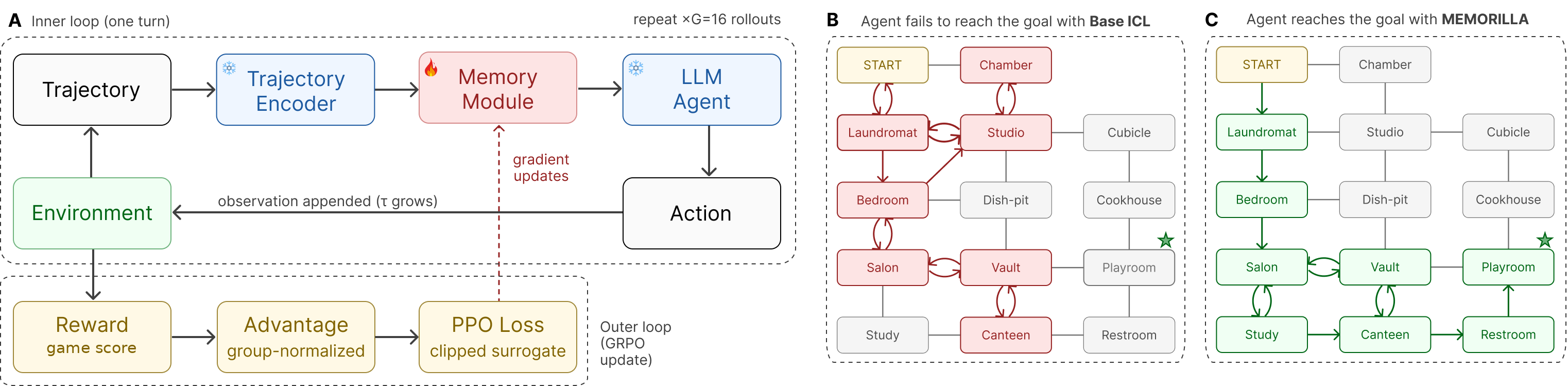}
\caption{\textbf{Reinforcement-learned memory on TextWorld.} \textbf{(a)} A frozen encoder embeds the trajectory, the trainable memory module compresses it to $K{=}8$ latents, and the frozen LLM agent acts; sparse rewards over $G{=}16$ rollouts drive a GRPO update to the memory module alone. \textbf{(b)} On a held-out game, ICL loops between explored rooms while \textbf{(c)} \memo reaches the goal (\textcolor{green!45!black}{$\star$}).}
\label{fig:textworld_obj}
\end{figure}

Beyond the supervised learning settings where the document collection can be static, we showcase the potential of \memo in scenarios where an LLM agent has to continuously learn and adapt from its environment. One such environment is TextWorld~\citep{cote2018textworld}, 
a suite of
partially observable text games in which an agent explores up to 65 rooms, manipulates objects and
pursues multi-step goals for at most 50 turns, with reward arriving mainly on a win.
After each turn,
\memo stores the completed turn as a new document and condenses the growing document embedding store into
$K=8$ memory tokens placed before the prompt
(Figure~\ref{fig:textworld_obj} and Appendix~\ref{app:textworld-rl}). 

\subsection{Experimental Setup}
\label{sec:textworld-setup}

The agent is a frozen Qwen3-4B-Instruct-2507 decoder paired with the frozen Qwen3-Embedding-4B encoder.
Only the memory module is trained, from random initialization and with GRPO on environment
reward (see Appendix~\ref{sec:rl_objective}). We compare against two baselines. Full-history in-context learning (ICL)
gives the decoder the complete interaction history without memory, and an untrained memory module
isolates the contribution of RL. Unlike in the supervised setting, the memory supplements the text
rather than replacing it. Every condition keeps the full history under the same input budget (16,000 tokens in training, 4,096 in evaluation), so any
improvement reflects what the memory learns to extract, not access to additional information. In the
medium and hard games, the history grows from 800 tokens at the first turn to about 3,200 and then
levels off, since the median rollout ends at turn 33 and only 5 of 40 rollouts reach turn 50.

\subsection{Results}



\begin{wrapfigure}{r}{2.6in}
\vspace{-9mm}
\centering
\includegraphics{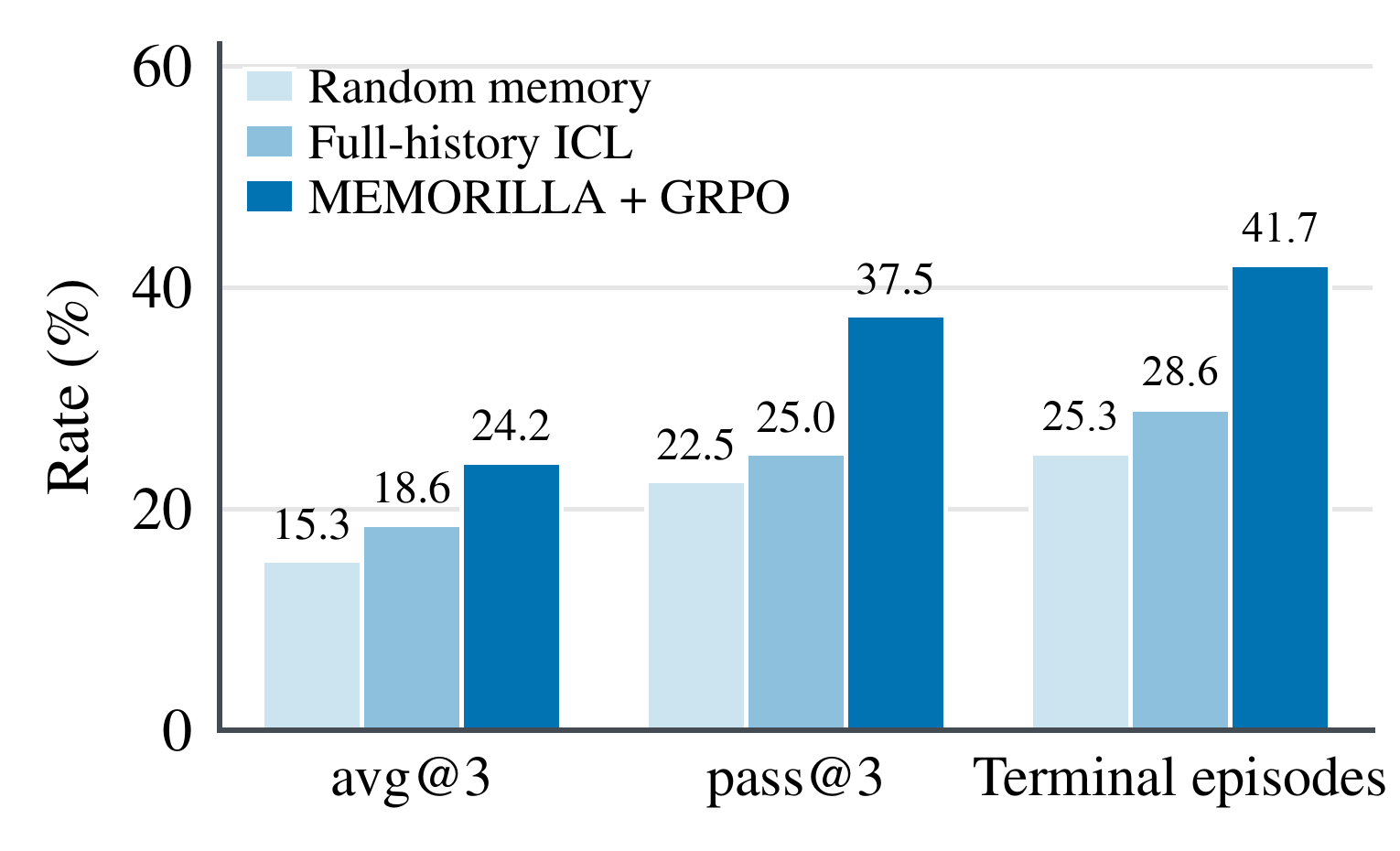}  
\caption{\textbf{\memo on TextWorld} with a frozen Qwen3-4B-Instruct-2507 decoder. 
Terminal episodes end through the game or the turn limit before the history exceeds the 4,096-token evaluation budget.}
\label{fig:textworld}
\vspace{-9mm}
\end{wrapfigure}

\paragraph{Task completion.} Training \memo with GRPO improves avg@3 from 0.186 to 0.242 and pass@3 from 0.250 to 0.375 relative to full-history ICL (Figure~\ref{fig:textworld}). Untrained memory instead falls below ICL on both metrics, avg@3 (0.153) and pass@3 (0.225), so the gain comes from learning rather than from the added tokens. The
trained memory acts as a compact state that keeps the frozen agent oriented over long trajectories.



\textbf{\memo reduces agent loops.} The largest behavioral change is termination rather than solve rate. ICL exhausts its 4,096-token budget on 71\% of rollouts against 58\% for \memo despite both reading the same trajectory text. Trajectory inspection shows that trained memory helps the agent escape navigation loops and execute keyed action sequences, such as unlocking and then opening a container. Figure~\ref{fig:textworld_obj} illustrates this, with ICL cycling between explored rooms and never reaching the goal, while \memo advances to the target room. The re-
maining failures concentrate in the longest games,
whose quests approach or exceed the 50-turn limit
and require backtracking across many rooms.

\section{Discussion and Conclusion}

\memo separates persistent knowledge storage from the context required for each query. New
documents can be added without updating the LLM, while the memory module learns to extract
relevant information through a fixed number of latent tokens. This design supports assistants that
draw on growing histories without proportionally increasing the decoder's context. Our analyses
suggest that the learned representations capture relationships across documents and recover personal
context beyond lexical similarity. In TextWorld, reinforcement learning improves a frozen agent's
decisions even when the interaction history is already available as text. Together, these findings
support learned memory as a means of improving how LLMs use accumulated information, with the
memory interface optimized through either answer supervision or environment reward.


\textbf{Conclusion.} In this work, we introduced \memo, a latent semantic memory system that gives frozen
LLMs access to query-dependent information from large corpora as `soft' memory tokens. 
\memo demonstrates that a frozen LLM can use accumulated knowledge through a learned, query-dependent interface, rather than relying only on retrieved text or an expanding prompt. \memo leads retrieval, compression and agent memory baselines on average at $20$ to
$4{,}212\times$ fewer document tokens across personalization, long-document QA, and knowledge-intensive
tasks. Additionally, \memo also enables continual learning in agentic RL, where
training only the memory module teaches an agent a new environment without overwriting prior
knowledge.


\textbf{Limitations and Future Work.} \memo can lose fine-grained factual details, likely because the document encoder was never trained for memory compression.
Future work exploring combinations of latent memory and text retrieval could combine the global structure that memory captures with the exact entities that retrieval recovers. Furthermore, encoders trained specifically for compression could close the remaining gap on fine-grained details.
Our experiments pair a Qwen encoder with a Qwen decoder, and testing how well the memory mechanism transfers across model families is also an interesting future area of research. Our evaluation covers personalization, long-document QA and knowledge-intensive QA, so \memo may not carry over directly to domains such as coding. Nonetheless, for personalization tasks, because the persistent memory store holds only embeddings, a multimodal encoder could extend the same interface to images, audio and video, giving assistants a memory of what a user has seen and heard. Since each user's memory is a separate store, it can be kept on device or deleted on request, which eases privacy concerns.
 Finally, 
because LLM decoding dominates latency, the efficiency gains show up mainly for large collections. Hence one can explore extensions of \memo where memory modules could share a single persistent store and inject memory tokens across LLMs of varying sizes.

\subsection*{AI use statement}

We used coding agents to implement and run experiments, and minimally used generative AI tools to polish the writing of the paper. As part of our method, GPT-5 generated the PersonalizationV4 personas, GPT-5.1 generated the question categories, chat topics and questions, and GPT-5-mini generated the dialogues (Appendix~\ref{app:pv4-pipeline}). We did not use generative AI tools for any other part of this work. The authors reviewed all AI-assisted work and verified the code and experimental results produced with coding agents before reporting them. We take responsibility for the final content of this work, including text, claims and artifacts produced with the aid of generative AI.

\subsection*{Ethics statement}

\memo stores and recalls information from user conversations, which raises privacy concerns when deployed on real user data. Because the memory store is external to the LLM, a user's embeddings can be removed without retraining, but stored embeddings can still encode sensitive information and deployments should obtain user consent and protect them accordingly. PersonalizationV4 consists of synthetic users based on LLM-generated personas and contains no data from real users. All persona profiles follow a single machine template.

\subsection*{Reproducibility statement}

Section~\ref{sec:method} describes the \memo architecture, and Sections~\ref{sec:training_datasets} and~\ref{sec:training_objective} describe the training curriculum and objective. Appendix~\ref{app:datasets} lists the training data, and Appendix~\ref{app:hyperparameters} reports the model configuration, training hyperparameters, document chunking and evaluation splits. Appendix~\ref{app:baselines} describes the implementation of each baseline, and Appendix~\ref{app:benchmark_details} describes the benchmarks, the PersonalizationV4 data pipeline and the context budget of each method. Appendix~\ref{app:textworld-rl} gives the full TextWorld RL setup. All supervised evaluations use greedy decoding and a fixed random seed. We are releasing the code, the training and evaluation data with precomputed embeddings, and the PersonalizationV4 data on our \href{https://memory.stanford.edu/}{project website}.

\subsubsection*{Acknowledgments}
We thank Ben Geist, Charilaos Kanatsoulis, Jonas De Schouwer, and Michael Bereket for discussions and feedback on our manuscript, Shayan Talaei for discussions about memory, Victor Li for initial brainstorming of the project, and Olivia Markham for initial experimentation. This work used Marlowe~\citep{marlowe2025}, Stanford University's GPU-based Computational Instrument, supported by Stanford HAI and Stanford Research Computing. We also gratefully acknowledge the support of NSF under Nos. CCF-1918940 (Expeditions), DMS-2327709 (IHBEM), IIS-2403318 (III); NIH under No. 1U24NS146314-01, Gates Foundation, Stanford Data Applications Initiative, Wu Tsai Neurosciences Institute, Stanford Institute for Human-Centered AI, Chan Zuckerberg Initiative, Amazon, Genentech, SAP, and SCBX. The content is solely the responsibility of the authors and does not necessarily represent the official views of the funding entities.

\bibliographystyle{iclr2027_conference}
\bibliography{main}

\clearpage
\appendix

\section{Related Work}
\label{app:related_work}

\textbf{Textual memory.}
Textual memory keeps source records readable and retrieves relevant records
at inference.
RAG~\citep{lewis_perez_piktus_petroni_karpukhin_goyal_kuttler_lewis_yih_rocktaschel_et_al_2020}
and REALM~\citep{guu_lee_tung_pasupat_chang_2020} retrieve passages from a
document index. Systems built for longer histories retain the same basic
interface while organizing the records differently. MemGPT~\citep{packer2023memgpt}
moves information between context tiers, MemoryBank~\citep{zhong2024memorybank}
stores conversations and summaries, and Mem0~\citep{chhikara_khant_aryan_singh_yadav_2025}
and A-MEM~\citep{xu2025mem} maintain structured records across sessions.
These systems preserve provenance and exact details, but the LLM can reason
only over material that retrieval places in its context. Evidence outside
the selected set remains unavailable, and widening that set yields limited
returns. In-Context RALM~\citep{ram2023context} finds that most
question-answering gains come from the first one or two retrieved documents.
MemoRAG~\citep{qian2025memorag} uses compressed global memory to generate
retrieval clues, but its final generator still reads selected passages.
Textual memory is therefore effective when evidence can be localized, while
signals distributed across a collection remain constrained by document
selection. \memo instead computes a query-conditioned latent state from the
full collection, complementing retrieval when exact evidence is needed.

\textbf{Latent compression.}
Latent compression replaces replayed text with compact continuous state.
ICAE~\citep{ge2024incontext} and IC-Former~\citep{wang2024context} encode a
long input into reusable vectors consumed by frozen target LLMs, while
AutoCompressors~\citep{chevalier2023adapting} fine-tune an LLM to produce and
consume recursive summary vectors. These representations can be cached
because they are computed without a query, but they cannot change what they
emphasize for different queries. Gist tokens~\citep{mu2023learning} and
500xCompressor~\citep{li2025500xcompressor} make the same tradeoff using
layer-specific key--value states rather than standard input embeddings.
xRAG~\citep{cheng2024xrag} makes the representation query relevant by first
selecting a document through retrieval and mapping its embedding to one LLM
token, which retains document selection as the access mechanism. A second
line improves the alignment between compression and generation.
PISCO~\citep{louis2025pisco} and COCOM~\citep{rau2024context} jointly adapt
their compressors and generators, while CLaRa~\citep{he2025clara} learns
reusable document vectors before training a query reasoner with the
generator. These methods gain a learned latent interface but require an
adapted generator. Other methods obtain reuse within a narrower memory
domain. TokMem~\citep{wu2026tokmem} optimizes a separate token for each
procedure, while ComMer~\citep{zeldes2025commer},
USER-LLM~\citep{ning2025user}, and
Embedding-to-Prefix~\citep{huber2025embedding} construct latent state
for user histories. \memo instead accepts unseen document collections
without further optimization and converts an LLM-independent store into new
query-conditioned input embeddings for a frozen LLM.

\textbf{Model memory.}
Model memory stores information in architecture-specific state.
MemoryLLM~\citep{wang2024memoryllm} maintains a self-updating latent pool,
M+~\citep{wang2025m} retrieves from long-term storage using the current
hidden state, and LongMem~\citep{wang2023augmenting} retrieves cached key--value
states through a trained side network. Their shared internal representation
supports query-conditioned recall, but also couples the memory and its
reader to a particular model. Other methods preserve state during model
execution rather than across an independent document collection.
LM2~\citep{kang2025lm2} updates a gated memory matrix during autoregressive
inference. Titans~\citep{behrouz2025titans} combines fixed task-level
parameters with a contextual neural memory updated while processing a
sequence. MELODI~\citep{chen2025melodi} and Activation
Beacon~\citep{zhang2025long} compress state across context windows and
transformer layers to extend long-sequence processing.
Cartridges~\citep{eyuboglu2026cartridges} makes model-specific key--value
caches reusable and composable, but requires offline optimization for each
new corpus. These approaches demonstrate the value of persistent latent
state, while their architecture-specific interfaces limit reuse across
models and collections. \memo instead keeps memory external and supplies it
to a frozen LLM as input embeddings.

\section{Additional Details on Training Methodology}

\subsection{Training Datasets}
\label{app:datasets}

Table~\ref{tab:training-data} details the training data used across the first two stages of the \memo pipeline. The pretraining corpus consists of approximately 2M paraphrase reconstruction examples derived from English Wikipedia articles. The finetuning mixture aggregates 1.34M queries across 10 sources spanning reading comprehension, open-domain QA, and summarization, exposing the memory module to varied compression demands across different answer types and document structures. PubMedQA contributes only its 900-question training split, which is disjoint from the 100 test questions used for evaluation.

\begin{table*}[h!]
\caption{\textbf{Training datasets for \memo.} The pretraining corpus is derived from English Wikipedia. The finetuning mixture spans reading comprehension, open-domain question answering, and summarization tasks across 10 sources. Sizes are numbers of training queries.}
\label{tab:training-data}
\centering
\scriptsize
\setlength{\tabcolsep}{3pt}
\begin{tabular*}{\textwidth}{@{\extracolsep{\fill}} l l l r @{}}
    \toprule
    \textbf{Stage} & \textbf{Source} & \textbf{Category} & \textbf{Num. Queries} \\
    \midrule
    \textbf{Pretraining} & EnWiki Articles & Paraphrase & 1,979,809 \\
    \cmidrule{1-4}
    \multirow[t]{11}{*}{\textbf{Finetuning}}
        & SQuAD v2 \citep{rajpurkar_2018_squad_v2}      & Reading Comprehension & 86,821 \\
        & DROP \citep{dua_2019_drop}                    & Reading Comprehension & 77,400 \\
        & CoQA \citep{reddy_2019_coqa}                  & Reading Comprehension & 108,647 \\
        & PubMedQA \citep{jin2019pubmedqadatasetbiomedicalresearch} & Reading Comprehension & 900 \\
        & QuAIL \citep{rogers_2020_quail}               & Reading Comprehension & 10,246 \\
        & PwC \citep{ge2024incontext}                   & Reading Comprehension & 241,564 \\
        & CNN/DailyMail \citep{see_2017_cnn_dm}         & Summarization         & 287,113 \\
        & SamSum \citep{gliwa_2019_samsum}              & Summarization         & 14,731 \\
        & DialogSum \citep{chen_2021_dialogsum}         & Summarization         & 12,460 \\
        & MS MARCO \citep{bajaj_2016_ms_marco}          & Open-domain QA        & 504,111 \\
    \cmidrule(l){2-4}
        & \multicolumn{2}{l}{\textbf{Finetuning total}} & \textbf{1,343,993} \\
    \bottomrule
\end{tabular*}
\end{table*}

\subsection{Training Configuration}
\label{app:hyperparameters}

All \memo runs use frozen encoder and decoder backbones. Only the memory module is trainable unless otherwise noted. Supervised training stages use bf16 mixed precision on 4 NVIDIA H100 80GB GPUs and random seed 23. Document embeddings are precomputed and cached to amortize encoding cost. Table~\ref{tab:chunking} lists how each benchmark's collection is split into the chunks $d_i$ that the encoder embeds. The supervised objective follows Section~\ref{sec:training_objective}, with loss computed only over answer tokens. A single memory module is trained jointly across all downstream tasks for up to 5 epochs. On PV4 and PMv2 we continue for additional epochs because both were still improving on validation when the other benchmarks had converged. Train and test users are disjoint on both, so one module serves every user and enrolling a new one requires only embedding their chats.

\begin{table}[h!]
\caption{\textbf{Document chunks per benchmark.} Each chunk $d_i$ is embedded once by the frozen encoder. Token counts use the Qwen3 tokenizer.}
\label{tab:chunking}
\centering
\scriptsize
\setlength{\tabcolsep}{4pt}
\begin{tabular*}{\textwidth}{@{\extracolsep{\fill}} l l r @{}}
    \toprule
    \textbf{Benchmark} & \textbf{Chunk unit} & \textbf{Tokens per chunk (mean / median)} \\
    \midrule
    PMv2        & One conversation session & 280 / 262 \\
    PV4         & One chat dialogue & 252 / 224 \\
    PubMedQA    & One abstract section & 84 / 78 \\
    TriviaQA    & Evidence passage, split sentence-greedily at 128 tokens & 70 / 77 \\
    NarrativeQA & Story paragraph, split on sentences above ${\sim}$256 tokens & 94 / 35 \\
    FactKG      & One knowledge-graph triple & 13 / 13 \\
    LaMP-4      & One profile item & 47 / 44 \\
    LaMP-7      & One profile tweet & 25 / 25 \\
    \bottomrule
\end{tabular*}
\end{table}

\begin{table*}[h!]
\caption{\textbf{Model configuration for \memo.} The memory module maps document representations into a fixed set of learned memory tokens that are inserted into the frozen decoder context.}
\label{tab:training-config-model}
\centering
\scriptsize
\setlength{\tabcolsep}{3pt}
\renewcommand{\arraystretch}{1.10}
\begin{tabular*}{\textwidth}{@{\extracolsep{\fill}} l l p{0.60\textwidth} @{}}
    \toprule
    \textbf{Component} & \textbf{Setting} & \textbf{Value} \\
    \midrule
    \multirow[t]{9}{*}{\textbf{Memory module}}
        & Pooling mode & Adaptive cross-attention with softmax over documents \\
        & Memory slots & $K{=}16$ \\
        & Cross-attention layers & 2 layers, slots $\leftarrow$ documents \\
        & Attention heads & 8 \\
        & Encoder-side dimension & 2560 \\
        & Decoder-side dimension & 2560 for 4B decoder, 4096 for 8B decoder \\
        & Slot initialization & Xavier uniform \\
        & Projection & Linear encoder-to-decoder projection \\
        & Regularization & Dropout 0.1 with residual cross-attention \\
    \cmidrule{1-3}
    \multirow[t]{4}{*}{\textbf{Frozen backbones}}
        & Document encoder & Qwen3-Embedding-4B, mean-pooled per document \\
        & 4B decoder & Qwen3-4B-Instruct-2507 \\
        & 8B decoder & Qwen3-8B \\
        & Trainable parameters & Memory module only; 190{,}190{,}080 for 4B and 194{,}123{,}776 for 8B \\
    \cmidrule{1-3}
    \multirow[t]{3}{*}{\textbf{Generation}}
        & Decoding & Greedy decoding with $T{=}0$, top-$p{=}1.0$, top-$k{=}{-}1$ \\
        & Short-answer limit & 8 tokens for classification, 64 tokens for QA and fact-verification tasks \\
        & Long-form limit & 128 tokens for LaMP-4 and LaMP-7; 512 tokens for EnWiki reconstruction \\
    \bottomrule
\end{tabular*}
\renewcommand{\arraystretch}{1.0}
\end{table*}

\begin{table*}[t!]
\caption{\textbf{Training stages for \memo.} Stage 1 pretrains memory on EnWiki paraphrase reconstruction, Stage 2 warm-starts on the mixed supervision corpus in Table~\ref{tab:training-data}, and Stage 3 performs downstream multi-task fine-tuning.}
\label{tab:training-config-stages}
\centering
\scriptsize
\setlength{\tabcolsep}{3pt}
\renewcommand{\arraystretch}{1.12}
\begin{tabular*}{\textwidth}{@{\extracolsep{\fill}} l p{0.22\textwidth} p{0.22\textwidth} p{0.28\textwidth} @{}}
    \toprule
    \textbf{Setting} & \textbf{Stage 1} & \textbf{Stage 2} & \textbf{Stage 3} \\
    & \textbf{EnWiki pretrain} & \textbf{Multi-task pretrain} & \textbf{Downstream fine-tune} \\
    \midrule
    Initialization & Random & Warm-start from Stage 1 & Warm-start from Stage 2 \\
    Data & EnWiki paraphrase reconstruction & Mixed supervision corpus & Unified downstream mixture \\
    Tasks & Paraphrase reconstruction & Reading comprehension, open-domain QA, summarization & PV4, PMv2, FactKG, TriviaQA, NQA, PubMedQA, LaMP-4, LaMP-7 \\
    Optimizer & AdamW & AdamW & AdamW \\
    LR scheduler & Cosine annealing with warmup & Cosine annealing with warmup & Cosine annealing with warmup \\
    Peak learning rate & $3{\times}10^{-4}$ & $3{\times}10^{-4}$ & $1{\times}10^{-4}$ \\
    Weight decay & $1{\times}10^{-2}$ & $1{\times}10^{-2}$ & $1{\times}10^{-2}$ \\
    Warmup ratio & 0.03 & 0.03 & 0.1 \\
    Max decoder length & 1024 & 1024 & 2048 \\
    Per-device batch size & 8 for 4B, 4 for 8B & 8 for 4B, 4 for 8B & 2 \\
    Gradient accumulation & 4 for 4B, 8 for 8B & 4 for 4B, 8 for 8B & 8 \\
    Effective batch size & 128 & 128 & 64 \\
    Devices & 4 GPUs & 4 GPUs & 4 GPUs \\
    Epochs & 1 & 1 & 5 \\
    Checkpoint selection & Final checkpoint & Final checkpoint & One shared multi-task checkpoint; best by validation loss only for PV4/PMv2 specialization \\
    Validation interval & --- & --- & 0.25 epoch \\
    Random seed & 23 & 23 & 23 \\
    \bottomrule
\end{tabular*}
\renewcommand{\arraystretch}{1.0}
\end{table*}

\begin{table*}[t!]
\caption{\textbf{Downstream training and evaluation details.} Stage 3 trains on a unified downstream mixture. PV4 and PMv2 receive additional benchmark-specific fine-tuning, while all other benchmarks use the multi-task checkpoint directly.}
\label{tab:training-config-eval}
\centering
\scriptsize
\setlength{\tabcolsep}{3pt}
\renewcommand{\arraystretch}{1.10}
\begin{tabular*}{\textwidth}{@{\extracolsep{\fill}} l p{0.76\textwidth} @{}}
    \toprule
    \textbf{Item} & \textbf{Configuration} \\
    \midrule
    Stage 1 corpus & 1.98M EnWiki paraphrase reconstruction examples, chunked at approximately 128 tokens with mean 1.73 documents per row \\
    Stage 2 corpus & Mixed supervision over SQuAD v2, DROP, CoQA, PubMedQA, QuAIL, PwC, CNN/DailyMail, SAMSum, DialogSum, and MS MARCO \\
    Stage 3 corpus & Unified concatenation of PV4, PMv2, FactKG, TriviaQA, NarrativeQA, PubMedQA, LaMP-4, and LaMP-7 \\
    PV4 and PMv2 specialization & After multi-task fine-tuning, PV4 and PMv2 are each further fine-tuned for 5 epochs on the corresponding benchmark training split using the same recipe \\
    Other downstream tasks & FactKG, TriviaQA, NarrativeQA, PubMedQA, LaMP-4, and LaMP-7 use the multi-task checkpoint directly with no additional task-specific training \\
    Evaluation split & Official test split for every task except LaMP-4 and LaMP-7, which use their development splits; TriviaQA, which uses a deterministic seed-23 validation subsample with $n{=}500$; and PubMedQA, which uses the 100-example test portion of a 900/100 split derived from PQA-L \\
    \bottomrule
\end{tabular*}
\renewcommand{\arraystretch}{1.0}
\end{table*}

\section{Additional Details on Baselines}
\label{app:baselines}

All baselines use the frozen Qwen3-8B as the decoder and the frozen Qwen3-Embedding-4B as the document encoder, and every answer is generated greedily. Baselines that insert text share one prompt template per benchmark and differ only in the context placed before the question.

\textbf{Closed-book.} The decoder receives only the query, which shows how much of each benchmark it answers from parametric knowledge.

\textbf{RAG and Full Context.} RAG-$k$ ranks the chunks of a collection by the cosine similarity between their embeddings and the query embedding, and places the top $k$ chunks in the prompt as text, most similar first and separated by blank lines. Full Context places every chunk in its stored order. Inputs longer than the enforced 32{,}768-token limit use YaRN with a scaling factor of 4, which extends the window to 131{,}072 tokens and applies to part of PMv2, TriviaQA and NarrativeQA and to all of PV4 (Table~\ref{tab:context-budget}). For the 10.8\% of evaluated NarrativeQA questions whose books exceed this window, the earliest chunks are dropped.

\textbf{LLMLingua-2.} LLMLingua-2~\citep{llmlingua2} is a hard prompt compression method that deletes tokens from the context rather than replacing it with latent vectors. We use the released XLM-RoBERTa-large token classifier. Because the classifier never sees the question, each collection is compressed once and reused for every query, similar in principle to \memo. We concatenate the documents of the Full Context input with blank-line separators and compress them as one block, retaining newlines and question marks. The question, answer-format instruction, system prompt and chat template are copied verbatim, so the prompt differs from Full Context only in the compressed documents. We disable the option to discard whole documents, since document selection is retrieval and is already covered by RAG-$k$. We set absolute budgets of 32, 64, 128 and 350 tokens rather than a compression rate, which the method treats only as a soft target, and report the best budget for each metric.

\textbf{xRAG.} xRAG~\citep{cheng2024xrag} maps the embedding of the top retrieved chunk to a single soft token in the input space of the decoder through a trained bridge. We reimplement the bridge in our codebase as a layer normalization followed by a two-layer MLP with a GELU activation, which maps the 2{,}560-dimensional chunk embedding to the 4{,}096-dimensional decoder input space with 27{,}276{,}288 trainable parameters. The chunk is retrieved by cosine similarity with the shared encoder. Only the bridge is trained, from scratch on each benchmark's training split with the same answer-token loss as \memo. We train for 5 epochs with AdamW at a learning rate of $10^{-4}$, a weight decay of 0.01, a warmup ratio of 0.1, an effective batch size of 16, a maximum sequence length of 8{,}192 tokens and seed 23.

\textbf{ComMer.} ComMer~\citep{zeldes2025commer} compresses user-history documents into latent representations with a learned compressor, merges them and passes them to a frozen LLM for personalized generation. We compare against its reported LaMP-4 and LaMP-7 results in Table~\ref{tab:commer}.

\textbf{Mem0 and MemoRAG.} Mem0~\citep{chhikara_khant_aryan_singh_yadav_2025} extracts and consolidates memories at write time, while MemoRAG~\citep{qian2025memorag} forms clue queries from a global memory to guide retrieval. We run the official implementations with the settings in Table~\ref{tab:mem0-memorag-config}. Both use a frozen Qwen3-4B-Instruct-2507 to write their memory, as the fact extractor of Mem0 and as the memory model of MemoRAG. This matches the 4B scale of the Qwen3-Embedding-4B encoder with which \memo reads the same collections, so every method processes the documents with a 4B model and answers with the same frozen Qwen3-8B decoder. MemoRAG runs in its supported long-context memory mode, in which the memory model reads the uncompressed collection rather than a compressed version of it. Its retrieval uses each benchmark's own chunks and the shared encoder, so the clue mechanism is its only difference from RAG-$k$. We sweep the retrieval depth of Mem0 over $k \in \{10, 30, 50\}$ and of MemoRAG over its native setting and caps of 5 and 50 chunks, and report each at its best depth per metric.

\begin{table*}[h!]
\caption{\textbf{Mem0 and MemoRAG configuration.} Rows below the rule are shared with the RAG-$k$ baselines, so both methods differ from RAG only in the context given to the decoder.}
\label{tab:mem0-memorag-config}
\centering
\scriptsize
\setlength{\tabcolsep}{3pt}
\renewcommand{\arraystretch}{1.10}
\begin{tabular*}{\textwidth}{@{\extracolsep{\fill}} l p{0.39\textwidth} p{0.39\textwidth} @{}}
    \toprule
    \textbf{Setting} & \textbf{Mem0} & \textbf{MemoRAG} \\
    \midrule
    Implementation & \texttt{mem0ai} 0.1.118 with its stock prompts and ADD/UPDATE/DELETE/NONE loop, unmodified & Official repository with its prompts unchanged \\
    Memory-writing model & Frozen Qwen3-4B-Instruct-2507 extracts facts and consolidates memories & Frozen Qwen3-4B-Instruct-2507 reads the collection as its memory \\
    Memory-model decoding & Greedy, up to 2{,}048 tokens & Greedy, up to 128 tokens per clue call \\
    Write path & One \texttt{add()} per document, with each extracted fact reconciled against its 5 nearest stored memories & Collection read once per entity and cached, front-truncated only where it exceeds the context window (NarrativeQA) \\
    Memory scope & One Qdrant collection per user (PMv2, PV4), story (NarrativeQA) or question (others) & One memory per user, story or question \\
    Retrieval query & The question & Clue spans and surrogate questions longer than 3 words, plus the question \\
    Retrieval & Top-$k$ memories by cosine similarity & Top 3 chunks per query, unioned, deduplicated and kept in corpus order \\
    Depths swept & $k \in \{10, 30, 50\}$ memories & Native (about 10 to 20 chunks), and capped at 5 and 50 by maximum similarity over queries \\
    Memory-model answer & Not applicable & Not used (\texttt{use\_memory\_answer=False}) \\
    \cmidrule{1-3}
    Retriever & \multicolumn{2}{p{0.79\textwidth}}{Qwen3-Embedding-4B with 2{,}560 dimensions and no instruction prefix, over each benchmark's chunks (Table~\ref{tab:chunking})} \\
    Generator & \multicolumn{2}{p{0.79\textwidth}}{Frozen Qwen3-8B with the system prompt, chat template, answer-format instruction and maximum new tokens of RAG-$k$} \\
    Decoding & \multicolumn{2}{p{0.79\textwidth}}{Greedy, with $T{=}0$, top-$p{=}1.0$ and top-$k{=}{-}1$} \\
    \bottomrule
\end{tabular*}
\renewcommand{\arraystretch}{1.0}
\end{table*}

\section{Additional Details on Benchmarks}
\label{app:benchmark_details}

\textbf{Personalization.}
These benchmarks require combining information spread across a user's past chats or writing into a picture of the user that no single document provides.

\begin{itemize}[leftmargin=8pt, topsep=0pt]
\item \textbf{PersonaMem-v2} (PMv2, $n{=}5{,}000$) \citep{jiang_yuan_shen_hao_xu_chen_liu_vijjini_he_yu_et_al_2025}. Each user has a long history of conversations with an assistant about everyday tasks, and each question asks for a response that fits the user's preferences. Most preferences are never stated outright but surface indirectly, for instance in an email the user asks to have polished, so the model must infer them from scattered cues across many sessions rather than look them up.
\item \textbf{PersonalizationV4} (PV4, $n{=}4{,}237$). Our synthetic benchmark gives each user 200 two-turn conversations that reflect a persona without stating its traits (Appendix~\ref{app:pv4-pipeline}). Each question places the user in a new scenario and requires combining at least two persona facts, which may appear in different conversations. A hard subset ($n{=}629$) keeps the questions that the untrained Qwen3-4B-Instruct-2507 decoder answers incorrectly without documents or with the top retrieved conversation.
\item \textbf{LaMP-4} ($n{=}1{,}925$), \textbf{LaMP-7} ($n{=}1{,}500$) \citep{salemi_mysore_bendersky_zamani_2023}. LaMP-4 asks the model to write a headline for a new news article in the style of the author's earlier articles and headlines, and LaMP-7 asks it to rewrite a tweet in the style of the user's earlier tweets. The challenge is stylistic rather than factual, since style emerges from the whole writing history rather than a single item.
\end{itemize}

\textbf{Long-Document QA.}
This axis tests whether memory can compress entire long documents into a usable representation for downstream QA.

\begin{itemize}[leftmargin=8pt, topsep=0pt]
\item \textbf{NarrativeQA} (NQA, $n{=}10{,}557$) \citep{kovcisky2018narrativeqa}. Questions concern full books and movie scripts and were written from plot summaries without access to the stories, so they target events, motives and relations described across the narrative rather than sentences that can be matched locally. Answers are free-form text, which rewards understanding the plot over copying a passage.
\item \textbf{TriviaQA} ($n{=}500$ from the $17{,}944$-question validation split) \citep{joshi_choi_weld_zettlemoyer_2017}. Questions written by trivia enthusiasts are paired with Wikipedia and web documents collected afterwards. Since the questions were not written against these documents, their wording and structure often differ from the passages that answer them, and an answer may require combining facts across sentences.
\end{itemize}

\textbf{Knowledge-Intensive QA.}
These benchmarks test factual grounding from structured or semi-structured evidence of varying quantity and format.

\begin{itemize}[leftmargin=8pt, topsep=0pt]
\item \textbf{FactKG} ($n{=}9{,}041$) \citep{kim_park_kwon_jo_thorne_choi_2023}. Each claim must be labeled as supported or refuted by facts from the DBpedia knowledge graph, through one-hop lookups, conjunctions of several facts, existence checks, multi-hop chains or negations. A multi-hop claim such as ``AIDAstella was built by a company in Papenburg'' requires chaining two facts, neither of which verifies the claim on its own, and a negated claim requires confirming that a relation is absent.
\item \textbf{PubMedQA} ($n{=}100$) \citep{jin2019pubmedqadatasetbiomedicalresearch}. Each question is a biomedical research question based on an article title, answered yes, no or maybe from the article's abstract with its conclusion removed. The model must infer the conclusion from the reported methods and results, often by interpreting quantitative findings. The entire source fits easily in the context, which makes PubMedQA a contrast case where condensing the source offers little benefit.
\end{itemize}

Table~\ref{tab:datasets} summarizes the benchmark datasets evaluated in this work. All datasets use standard train/validation/test splits. For the personalization benchmarks (PV4, PMv2, LaMP-4, LaMP-7), training and evaluation users are strictly disjoint to prevent leakage of user-specific information. We evaluate LaMP-4 and LaMP-7 on their development splits and TriviaQA on the validation split. For PubMedQA we use a 900/100 train/test split derived from the PQA-L source. Table~\ref{tab:context-budget} reports how many document tokens each method places in the decoder per query. The serving stack enforces a 32{,}768-token decoder limit and enables YaRN for every full-context example that exceeds it, which covers part of PMv2, TriviaQA and NarrativeQA and all of PV4. Every PubMedQA collection and 96.7\% of LaMP-7 collections hold at most 50 chunks, so RAG-50 reads the same chunks as Full Context on them.

\begin{table}[h!]
\caption{\textbf{Benchmark datasets overview.} Each row corresponds to an evaluation benchmark. Each data point consists of a textual query paired with a collection of source documents from which the model must produce an answer or generation. $\bar{\mu}$ docs/q and $\bar{\mu}$ tok/q report the mean number of source documents and total source tokens per query in the test split.}
\label{tab:datasets}
\centering
\scriptsize
\setlength{\tabcolsep}{1pt}
\begin{tabular*}{\textwidth}{@{\extracolsep{\fill}}>{\raggedright\arraybackslash}p{0.35\textwidth}>{\raggedright\arraybackslash}p{0.25\textwidth}rrrrr@{}}
    \toprule
    \textbf{Dataset} & \textbf{Task} & $\bar{\mu}$ \textbf{docs/q} & $\bar{\mu}$ \textbf{tok/q} & \textbf{Train} & \textbf{Val} & \textbf{Test} \\
    \midrule
    PersonalizationV4 (PV4) (Ours)   & Personalization SA        & 200      & 50{,}476  & 15{,}058 & 1{,}755 & 4{,}237 \\
    PersonaMem-v2 (PMv2) \citep{jiang_yuan_shen_hao_xu_chen_liu_vijjini_he_yu_et_al_2025} & Personalization SA & 116 & 31{,}618 & 18{,}521 & 2{,}059 & 5{,}000 \\
    NarrativeQA (NQA) \citep{kovcisky2018narrativeqa} & Long-document generative QA & 1{,}420 & 76{,}225 & 32{,}747 & 3{,}461 & 10{,}557 \\
    FactKG \citep{kim_park_kwon_jo_thorne_choi_2023} & KG fact verification & 129 & 1{,}665 & 86{,}367 & 13{,}266 & 9{,}041 \\
    TriviaQA \citep{joshi_choi_weld_zettlemoyer_2017} & Open-domain factoid QA & 164 & 11{,}861 & 138{,}384 & 17{,}944 & --- \\
    PubMedQA \citep{jin2019pubmedqadatasetbiomedicalresearch} & Biomedical QA & 4 & 324 & 900 & --- & 100 \\
    LaMP-4 \citep{salemi_mysore_bendersky_zamani_2023} & Personalized headline gen. & 172 & 7{,}638 & 12{,}527 & --- & 1{,}925 \\
    LaMP-7 \citep{salemi_mysore_bendersky_zamani_2023} & Personalized tweet paraphrase & 18 & 465 & 10{,}437 & --- & 1{,}500 \\
    \bottomrule
\end{tabular*}
\end{table}

\subsection{PersonalizationV4 Data Pipeline}
\label{app:pv4-pipeline}

PersonalizationV4 is a synthetic benchmark for evaluating multi-document user modeling. Each user has one LLM-generated persona profile seeded from the Faithful Persona-based Conversational Dataset~\citep{jandaghi2023faithfulpersonabasedconversationaldataset}. The resulting profiles are 10k to 22k characters long (median 15.5k) and describe demographics, personality, work, hobbies, communication style and how preferences change over time. All of the user's data is derived from this profile through the five steps below, and Table~\ref{tab:pv4-example} traces one held-out user through each of them.

\begin{enumerate}[leftmargin=12pt, topsep=0pt]
\item \textbf{Question categories.} GPT-5.1 reads the persona and extracts 10 persona-specific categories that questions can test.

\item \textbf{Chat topics.} GPT-5.1 writes exactly 200 one-sentence scenarios per user in which the user asks the assistant for help, covering every part of the persona.

\item \textbf{Dialogues.} GPT-5-mini expands each topic into one two-turn dialogue with one user turn and one assistant turn. The user embodies the persona implicitly without stating its traits. Exactly 200 dialogues are generated per user and all form the collection that the memory module reads; there is no truncation step.

\item \textbf{Questions.} For each category, GPT-5.1 is asked to write 15 questions from the persona alone. It writes 14 in 1{,}234 of the 1{,}490 released category files and 13 to 15 in all but one, which has 5. Each question has a reference answer and four distractors. The instructions in Table~\ref{tab:pv4-prompt} require every question to combine at least two persona facts in a new scenario and every distractor to be a plausible preference that matches the reference answer in length, structure and specificity. This gives 133 to 145 questions per user (mean 141.3).

\item \textbf{Randomization.} The reference answer's position among its distractors is shuffled with a fixed seed of 42.
\end{enumerate}

Questions are posed without their candidate answers, and the model answers in free text. The reference answer and distractors are used only by the short-answer scorer, which counts a response as correct when its nearest candidate under the Qwen3-Embedding-4B encoder is the reference answer. We release 149 users: 119 training users and 30 held-out evaluation users, none of whom appear in training. One training user whose generated persona was a refusal is withheld from the release. Every question for these held-out users is evaluated: 139 to 143 per user and 4{,}237 in total, all in the test split. The PersonalizationV4 dataset is linked on our \href{https://memory.stanford.edu/}{project website}. The hard subset ($n{=}629$) is defined from the untrained Qwen3-4B-Instruct-2507 decoder's multiple-choice answers and contains questions it answers incorrectly without documents, with the top-1 retrieved document, or in both settings; 299 of the 629 questions are answered incorrectly in both settings.

\begin{table}[h!]
\caption{\textbf{One held-out PersonalizationV4 user through the pipeline.} Excerpts are verbatim apart from abridgement. The question draws on base-building, which appears only in the dialogue shown, and on the Iron Heart project, which appears in 18 other dialogues, so answering requires combining documents. The distractors are never shown to the model and serve only as scoring candidates.}
\label{tab:pv4-example}
\centering
\scriptsize
\setlength{\tabcolsep}{4pt}
\renewcommand{\arraystretch}{1.10}
\begin{tabular*}{\textwidth}{@{\extracolsep{\fill}} l p{0.80\textwidth} @{}}
    \toprule
    \textbf{Persona} & Name Maya Patel. Age 34. Location Austin, Texas, USA. Occupation Certified Welder \& Part-Time CNC Operator at a regional aerospace supplier. \\
     & \ldots\ Maya grew up in a multicultural household where her Indian-American parents emphasized discipline, education, and the importance of physical well-being. After completing a two-year technical diploma in welding technology, she entered the workforce in a traditionally male-dominated industry, where she quickly earned a reputation for precision, reliability, and an unflappable calm under pressure. Though she never pursued a university degree, her practical knowledge of metallurgy, robotics-aided fabrication, and safety protocols rivals that of many engineers. \ldots \\
    \midrule
    \textbf{Categories} & Aerospace welding expertise \& mental rehearsal habits. Gaming genre preferences \& playstyle strategies. Morning mango ritual \& mobility routine. \ldots\ (3 of 10) \\
    \midrule
    \textbf{Chat topic} & Maya asks for a list of cooperative building-and-craft games she can play online with classmates after welding class. \\
    \midrule
    \textbf{User turn} & \ldots\ looking for coop building-and-craft games I can jump into with the welding class after tonight's session. Need stuff that's accessible for mixed skill levels, supports voice chat, and has meaningful base-building or automation mechanics. Toss me a short list with platform notes and why each one fits. \\
    \textbf{Assistant turn} & \emph{(seven games with platform notes, from Minecraft to Raft)} \ldots\ Each supports voice via Discord/Steam, scales with mixed experience, and offers clear building goals for group sessions. \\
    \midrule
    \textbf{Question} & In a new open-world crafting RPG, players can specialize in one mastery for unique bonuses. Considering Maya's welding expertise, love for base-building, and ergonomic focus from the Iron Heart project, which specialization best aligns with her likely in-game identity? \\
    \midrule
    \textbf{Reference answer} & Master Builder, optimizing structural stability, workstation placement, and production comfort, because she prefers engineering-style layouts that echo her fabrication and ergonomic design background. \\
    \midrule
    \textbf{Distractors} & Arcane Historian, optimizing lore discovery and dialogue outcomes, because she prefers uncovering story fragments over experimenting with materials and structural layouts. \\
     & Beast Whisperer, optimizing pet behavior and breeding trees, because she prefers managing biological companions over designing durable and efficient physical constructs. \\
     & Shadow Scavenger, optimizing stealth routes and loot efficiency, because she prefers sneaking past threats over methodically constructing and upgrading comprehensive bases. \\
     & Combat Tactician, optimizing battlefield formations and ambushes, because she prefers front-line tactics over refining structures, workstations, or long-term infrastructure planning. \\
    \bottomrule
\end{tabular*}
\renewcommand{\arraystretch}{1.0}
\end{table}

\begin{table}[h!]
\caption{\textbf{Question-generation instructions.} Verbatim excerpts from the prompt given to GPT-5.1 for each category, which also contains the full persona profile, the target category and the number of questions to write.}
\label{tab:pv4-prompt}
\centering
\scriptsize
\setlength{\tabcolsep}{4pt}
\renewcommand{\arraystretch}{1.10}
\begin{tabular*}{\textwidth}{@{\extracolsep{\fill}} l p{0.80\textwidth} @{}}
    \toprule
    \textbf{Questions} & Each question must require synthesizing information or understanding nuance from the persona. \\
     & Questions must require combining two or more facts from the persona profile. Avoid questions that can be answered by recalling a single fact. \\
     & Embed each question in a specific, realistic scenario that requires applying persona knowledge to a novel situation rather than simple recall. \\
     & The question stem must not hint at characteristics of the answer. \\
    \midrule
    \textbf{Distractors} & Every choice (A-E) must be a reasonable, professional, or logical preference for a human. Never use obviously wrong, absurd, or incompetent distractors. \\
     & Ensure distractors include industry-standard best practices that most people in the persona's role would choose, but which this specific persona rejects. \\
     & Use information that is true about the persona but applies it to the wrong situation or frequency. \\
     & All five choices must have the same word count within a tolerance of two words. \\
    \bottomrule
\end{tabular*}
\renewcommand{\arraystretch}{1.0}
\end{table}

\subsection{Context Budget per Query}
\label{app:context-budget}

Table~\ref{tab:context-budget} reports the number of document-derived tokens each method places in the decoder for one query. We exclude the question, the answer-format instruction, the system prompt and the chat template, which are identical across methods. For each text baseline, the count is the length of its prompt minus the length of the closed-book prompt for the same question, measured with the Qwen3-8B tokenizer and averaged over the evaluated questions. Closed-book receives no document tokens, \memo contributes its 16 memory tokens and xRAG a single token. Figure~\ref{fig:main_results} places each method at the geometric mean of these counts over the eight benchmarks, and the compression column of Table~\ref{tab:main-results} divides the Full Context count by the 16 tokens of \memo.

\begin{table}[h!]
\caption{\textbf{Document tokens per query.} Mean number of document-derived tokens in the decoder input, excluding the question and all shared instructions. Mem0, MemoRAG and LLMLingua-2 are counted at the setting reported for each benchmark's primary metric in Table~\ref{tab:main-results}, and truncated inputs are counted as read. Fits 32K reports whether every Full Context prompt fits the enforced 32{,}768-token limit, and Partial means that YaRN is enabled for the remaining examples. $^{\dagger}$ marks benchmarks where RAG-50 reads the same chunks as Full Context, for every PubMedQA question and 96.7\% of LaMP-7 questions.}
\label{tab:context-budget}
\centering
\scriptsize
\setlength{\tabcolsep}{2.5pt}
\begin{tabular*}{\textwidth}{@{\extracolsep{\fill}} l r c r r r r r r r r r @{}}
    \toprule
    \textbf{Benchmark} & \textbf{Docs/q} & \makecell{\textbf{Fits}\\\textbf{32K}} & \makecell{\textbf{RAG}\\\textbf{1}} & \makecell{\textbf{RAG}\\\textbf{5}} & \makecell{\textbf{RAG}\\\textbf{50}} & \makecell{\textbf{Full}\\\textbf{Context}} & \makecell{\textbf{LLM}\\\textbf{Lingua-2}} & \textbf{xRAG} & \textbf{Mem0} & \makecell{\textbf{Memo}\\\textbf{RAG}} & \textbf{\memo} \\
    \midrule
    PMv2        & 116     & Partial & 125 & 652     & 9{,}009  & 31{,}587 & 250 & 1 & 665 & 9{,}254  & 16 \\
    PV4         & 200     & No      & 201 & 1{,}021 & 10{,}832 & 50{,}455 & 180 & 1 & 978 & 11{,}046 & 16 \\
    PubMedQA    & 4       & Yes     & 69  & 321     & 324$^{\dagger}$ & 324 & 290 & 1 & 8 & 324 & 16 \\
    NarrativeQA & 1{,}420 & Partial & 69  & 354     & 3{,}612  & 67{,}385 & 596 & 1 & 493 & 800      & 16 \\
    TriviaQA    & 164     & Partial & 51  & 226     & 2{,}357  & 11{,}888 & 243 & 1 & 269 & 303      & 16 \\
    FactKG      & 129     & Yes     & 13  & 64      & 568      & 1{,}727  & 279 & 1 & 300 & 573      & 16 \\
    LaMP-4      & 172     & Yes     & 45  & 227     & 1{,}715  & 7{,}580  & 227 & 1 & 251 & 230      & 16 \\
    LaMP-7      & 18      & Yes     & 25  & 125     & 435$^{\dagger}$ & 459 & 299 & 1 & 81 & 135 & 16 \\
    \midrule
    Geometric mean & & & 55 & 273 & 1{,}808 & 6{,}301 & 278 & 1 & 212 & 784 & 16 \\
    \bottomrule
\end{tabular*}
\end{table}

\section{Additional Results and Ablations}
\label{app:full-supervised-results}

Table~\ref{tab:main-results} reports the complete 8B results underlying the main quantitative comparison in Figure~\ref{fig:main_results}. Table~\ref{tab:commer} compares \memo with ComMer on LaMP-4 and LaMP-7.

\begin{table}[h!]
\caption{\textbf{Main results across eight benchmarks.} All methods use a frozen Qwen3-Embedding-4B encoder and a frozen Qwen3-8B decoder; Mem0 and MemoRAG also use a frozen Qwen3-4B-Instruct-2507 to build their memories. RAG-$k$ retrieves the top-$k$ chunks by cosine similarity and Full Context reads the full collection. \memo uses $K{=}16$ memory tokens. Best result per row is \textbf{bold}, second best is \underline{underlined}. The Compression column reports how many times more document tokens Full Context gives the decoder than the 16 memory tokens of \memo (Table~\ref{tab:context-budget}). LLMLingua-2 was not evaluated on the PV4 hard subset.}
\label{tab:main-results}
\centering
\scriptsize
\setlength{\tabcolsep}{1.4pt}
\begin{tabular*}{\textwidth}{@{\extracolsep{\fill}} l l c c c c c c c c c c c @{}}
    \toprule
    \textbf{Task} & \textbf{Metric} & \makecell{Closed\\book} & \makecell{RAG\\1} & \makecell{RAG\\5} & \makecell{RAG\\50} & \makecell{Full\\Context} & \makecell{LLM\\Lingua-2} & xRAG & Mem0 & \makecell{Memo\\RAG} & \textbf{\memo} & Compression \\
    \midrule
    PMv2       & Acc (SA)  & 25.64 & 26.10 & 25.92 & 26.72 & 27.92 & 26.92 & \underline{30.90} & 27.48 & 28.76 & \textbf{44.60} & 1{,}974$\times$ \\
    \cmidrule{1-13}
    PV4        & Acc (SA)       & 58.30 & 62.21 & 64.98 & 65.78 & 59.12 & 52.14 & 62.64 & 65.07 & \textbf{67.88} & \underline{66.65} & 3{,}153$\times$ \\
               & Acc (SA, Hard) & 31.96 & 38.00 & 39.27 & 41.97 & 36.57 & ---   & 34.98 & \underline{44.20} & \textbf{45.95} & 41.97 & \\
    \cmidrule{1-13}
    PubMedQA   & Acc & 21.00 & 34.00 & \underline{67.00} & \underline{67.00} & \underline{67.00} & \textbf{69.00} & 36.00 & 24.00 & \underline{67.00} & 59.00 & 20$\times$ \\
    \cmidrule{1-13}
    NarrativeQA
               & BLEU-1   & 7.09  & 7.36  & 7.85  & 8.29  & 5.89  & 6.87  & \textbf{16.86} & 7.77  & 9.14  & \underline{16.36} & 4{,}212$\times$ \\
               & ROUGE-1  & 11.54 & 13.32 & 15.60 & 18.45 & 19.12 & 7.26  & 17.33 & 14.24 & \textbf{21.80} & \underline{19.68} & \\
               & ROUGE-L  & 10.71 & 12.44 & 14.59 & 17.23 & 17.99 & 7.18  & 16.92 & 13.80 & \textbf{20.51} & \underline{19.31} & \\
               & F1-Score & 9.87  & 10.51 & 11.64 & 13.27 & 13.50 & 5.76  & \underline{15.22} & 11.58 & 15.03 & \textbf{16.56} & \\
    \cmidrule{1-13}
    TriviaQA   & EM       & 36.00 & 51.80 & 55.20 & 55.60 & 52.60 & 42.40 & 47.00 & 44.60 & \underline{56.60} & \textbf{57.00} & 743$\times$ \\
               & BLEU-1   & 29.12 & 40.29 & 39.09 & 43.07 & 35.56 & 29.28 & \underline{46.54} & 29.09 & 45.00 & \textbf{58.66} & \\
               & BLEU-4   & 3.74  & 8.45  & 7.54  & 10.94 & 9.15  & 4.30  & \underline{11.31} & 5.54  & 10.35 & \textbf{14.79} & \\
               & F1-Score & 37.68 & 52.06 & 54.99 & \underline{58.05} & 56.49 & 39.74 & 50.37 & 42.74 & 57.55 & \textbf{60.60} & \\
    \cmidrule{1-13}
    FactKG     & Acc & 65.97 & 73.93 & 76.95 & 78.39 & \underline{80.43} & 71.65 & 78.98 & 72.79 & 80.38 & \textbf{84.33} & 108$\times$ \\
    \cmidrule{1-13}
    LaMP-4     & ROUGE-1  & 14.24 & 16.15 & 16.98 & 16.98 & 16.21 & 14.98 & \underline{18.12} & 14.72 & 17.08 & \textbf{20.30} & 474$\times$ \\
               & ROUGE-L  & 12.24 & 14.31 & 14.94 & 15.09 & 14.21 & 12.98 & \underline{16.10} & 12.70 & 15.02 & \textbf{18.15} & \\
               & F1-Score & 8.98  & 11.21 & 12.15 & 12.15 & 11.32 & 8.71  & \underline{13.46} & 9.05  & 12.41 & \textbf{15.54} & \\
               & METEOR   & 7.13  & 9.78  & 10.95 & 10.95 & 9.85  & 6.99  & 10.11 & 7.28  & \underline{11.21} & \textbf{11.41} & \\
    \cmidrule{1-13}
    LaMP-7     & ROUGE-1  & 48.73 & 49.80 & 49.80 & 49.65 & 49.57 & 49.61 & \underline{52.78} & 49.58 & 49.92 & \textbf{54.40} & 29$\times$ \\
               & ROUGE-L  & 43.13 & 44.11 & 43.81 & 43.77 & 43.71 & 43.73 & \underline{46.84} & 43.68 & 44.00 & \textbf{48.37} & \\
               & F1-Score & 36.10 & 37.28 & 37.28 & 37.26 & 37.14 & 37.17 & \underline{42.99} & 36.87 & 37.55 & \textbf{44.96} & \\
               & METEOR   & 32.31 & 33.70 & 33.78 & 33.73 & 33.55 & 33.74 & \underline{38.35} & 33.30 & 34.12 & \textbf{40.37} & \\
    \bottomrule
\end{tabular*}
\end{table}

\begin{table}[b!]
\centering
\caption{\textbf{Comparison with ComMer} on personalized generation, against ComMer's reported results. ComMer uses Gemma-2b while \memo uses Qwen3.}
\label{tab:commer}
\scriptsize
\setlength{\tabcolsep}{6pt}
\begin{tabular}{@{}lccc@{}}
    \toprule
    \textbf{Method} & \textbf{Tok.} & \textbf{LaMP-4} & \textbf{LaMP-7} \\
    & & ROUGE-L & ROUGE-L \\
    \midrule
    ComMer       & 128 & 17.50 & 45.00 \\
    \midrule
    \memo (4B)   & 16  & \underline{17.72} & \underline{47.04} \\
    \memo (8B)   & 16  & \textbf{18.15} & \textbf{48.37} \\
    \bottomrule
\end{tabular}
\end{table}


\subsection{Memory Architecture}

We port the BLIP-2 bridging module~\citep{blip2} to document collections and report it as a \textbf{Q-Former} architecture ablation.
Table~\ref{tab:architecture-ablation} compares methods for constructing 16 memory slots from $N$ document embeddings on PMv2. Each method is trained from scratch using the hyperparameters in Table~\ref{tab:training-config-stages}.

\textbf{Global mean-pool.} This baseline averages all document embeddings into one collection vector and copies it into all 16 slots.

\textbf{Segmented mean-pool.} This baseline partitions the ordered sequence of document embeddings into 16 adjacent groups of documents and averages each group to form one slot. The groups contain complete documents rather than subdivisions of individual documents.

\textbf{Segmented with matched parameters.} This method begins with the 16 slots produced by segmented mean-pooling and increases capacity to match \memo without learned queries or query conditioning.

\textbf{Cosine top-16 with matched parameters.} This method selects the 16 document embeddings with the highest cosine similarity to the question and uses them as the memory slots. Its slot processor matches the parameter budget of \memo.

\textbf{Documents only.} This Q-Former variant self-attends over its learned latent slots and then cross-attends to all document embeddings without access to the question.

\textbf{Q-Former.} This method self-attends over the learned latent slots and question embedding before the slots cross-attend to all document embeddings.

\textbf{\memo.} Our method first conditions the learned slots on the question embedding and then uses the conditioned slots to cross-attend to all document embeddings. It does not use self-attention.

Query-independent pooling is weakest and parameters do not fix it, since segmented pooling at matched budget reaches 31.10 on short-answer against 44.04. The Q-Former, which also conditions on the question, nearly matches \memo on PMv2 (43.34 against 44.04). On FactKG, \memo also outperforms the Q-Former (82.56 against 80.48), despite the Q-Former using 268{,}856{,}320 parameters. The documents-only variant reaches 78.72 on FactKG with 268{,}856{,}320 parameters. Replacing every document embedding with Gaussian noise changes the documents-only result by only +0.04 while lowering \memo by 5.93 and the Q-Former by 5.11. The gain therefore comes from conditioning the slots before they read rather than from either interaction on its own.

\begin{table}[h!]
\caption{\textbf{Memory architecture ablation.} All methods use the frozen Qwen3-Embedding-4B encoder and Qwen3-4B-Instruct-2507 decoder and share the training recipe, prompts, decoding and scorer. They are trained from scratch, so values are not comparable to Table~\ref{tab:main-results}. The Q-Former variants use more parameters than \memo{}.}
\label{tab:architecture-ablation}
\centering
\scriptsize
\setlength{\tabcolsep}{5pt}
\begin{tabular*}{\textwidth}{@{\extracolsep{\fill}} l l r c @{}}
    \toprule
    \textbf{Method} & \textbf{Query aware} & \textbf{Params} & \textbf{PMv2 SA} \\
    \midrule
    Global mean-pool                   & no           & 6{,}566{,}400   & 28.94 \\
    Segmented mean-pool                & no           & 6{,}566{,}400   & 29.68 \\
    Segmented, parameter-matched       & no           & 190{,}149{,}120 & 31.10 \\
    Cosine top-16, parameter-matched   & yes (hard)   & 190{,}149{,}120 & 33.86 \\
    \midrule
    Documents only & no          & 268{,}856{,}320 & 34.72 \\
    Q-Former                           & yes (soft)  & 268{,}856{,}320 & 43.34 \\
    \midrule
    \memo{}                            & yes (soft)   & 190{,}190{,}080 & \textbf{44.04} \\
    \bottomrule
\end{tabular*}
\end{table}

\subsection{Memory Capacity at Training Time}
\begin{figure}[h!]
\centering
\includegraphics{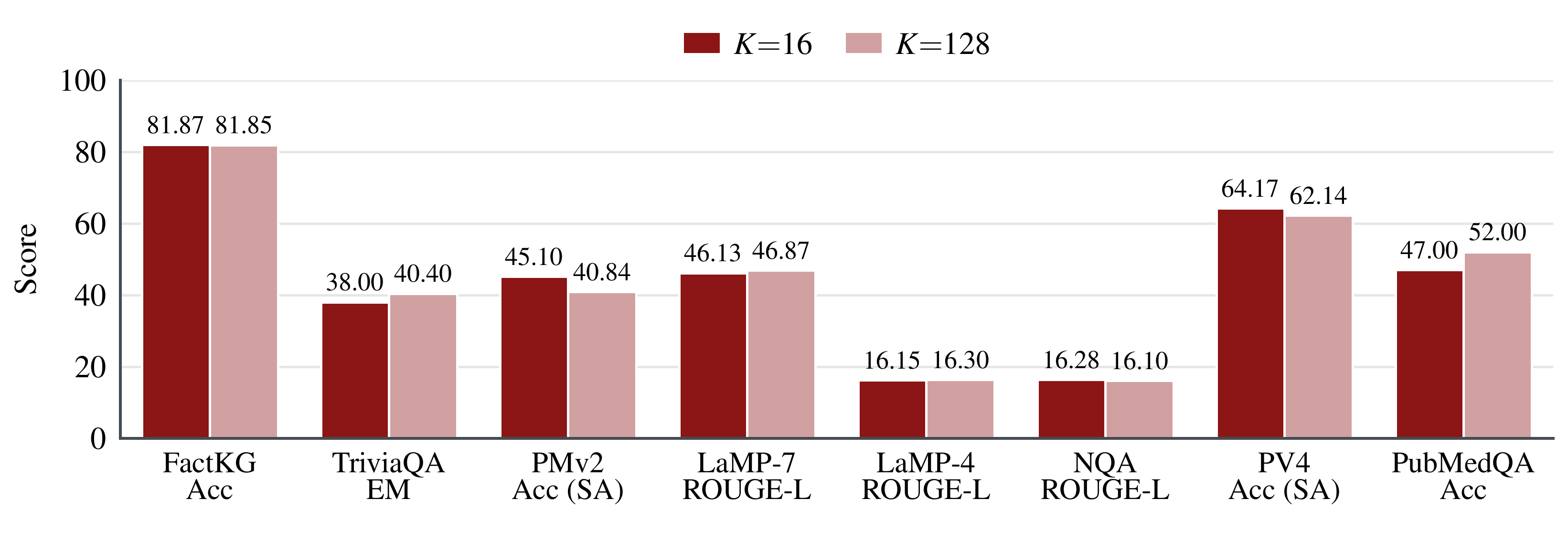}  
\caption{\textbf{Memory-token capacity.} All models use a Qwen3-4B-Instruct-2507 decoder and are trained from scratch under identical settings, differing only in the number of memory tokens $K$ ($K{=}16$ is the setting used throughout).}
\label{fig:k-training}
\end{figure}

Figure~\ref{fig:k-training} compares models trained from scratch at $K{=}16$ and $K{=}128$ under otherwise identical settings. All runs use the Qwen3-4B-Instruct-2507 decoder and the same hyperparameters reported in Table~\ref{tab:training-config-stages}, with $K$ as the only changed hyperparameter. Between $K{=}16$ and $K{=}128$, four of eight benchmarks move by less than a point at eight times the capacity and PMv2 and PV4 drop, so we set $K{=}16$ throughout. All runs are trained from scratch, so this concerns capacity rather than the deployed checkpoint.

\subsection{Memory Token Budgets for Inference}

\begin{figure}[h!]
\centering
\begin{subfigure}[t]{1.36in}\centering
  \includegraphics{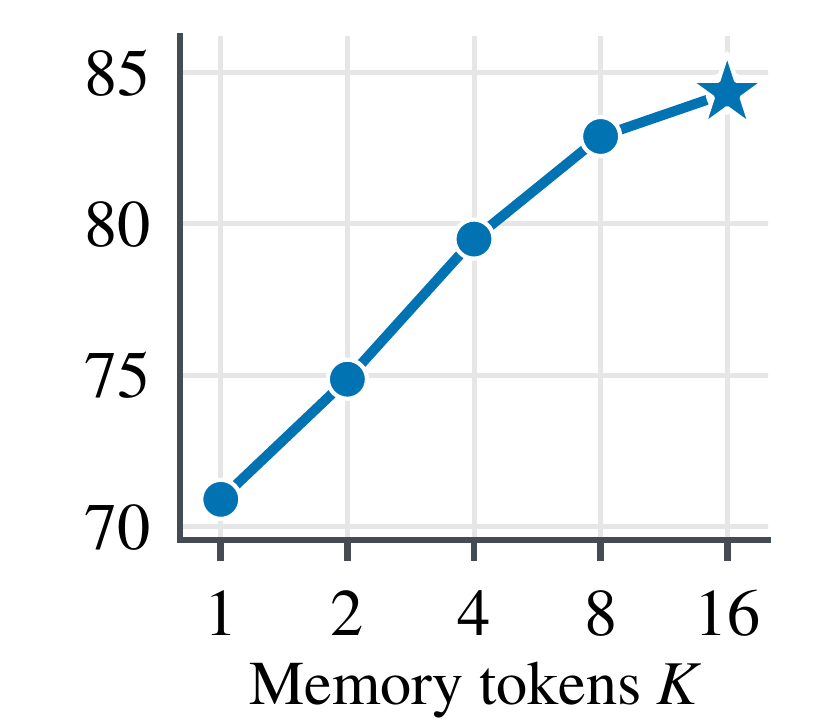}
  \caption{FactKG (Acc)}\label{fig:token-budget-factkg}
\end{subfigure}\hfill
\begin{subfigure}[t]{1.36in}\centering
  \includegraphics{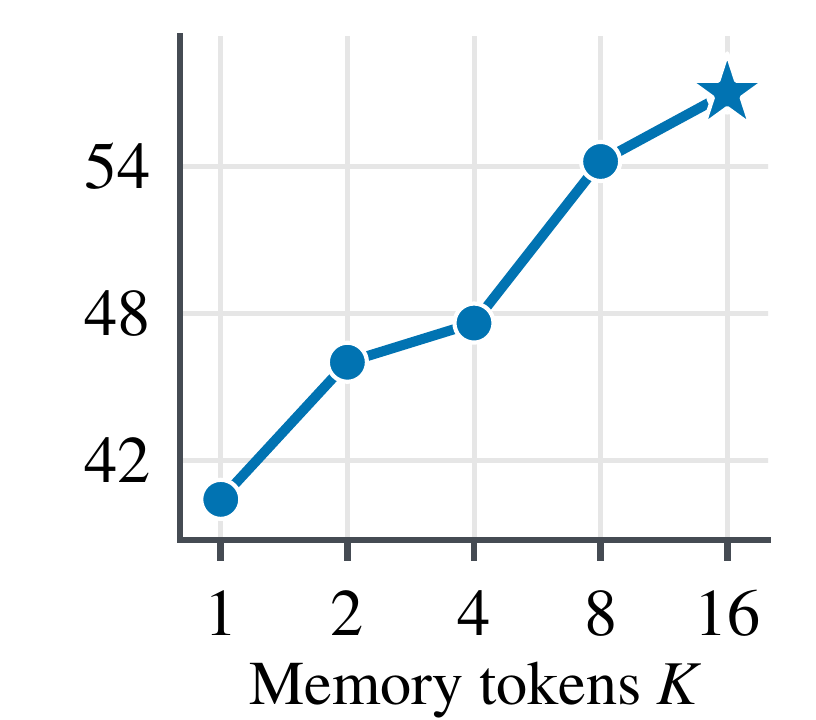}
  \caption{TriviaQA (EM)}\label{fig:token-budget-triviaqa}
\end{subfigure}\hfill
\begin{subfigure}[t]{1.36in}\centering
  \includegraphics{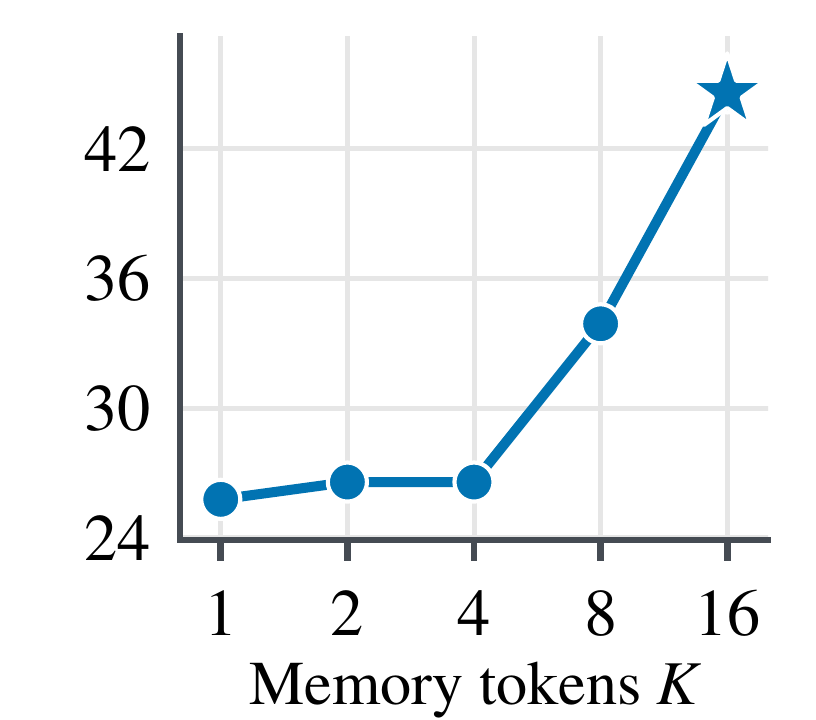}
  \caption{PMv2 (Acc)}\label{fig:token-budget-pmv2}
\end{subfigure}\hfill
\begin{subfigure}[t]{1.36in}\centering
  \includegraphics{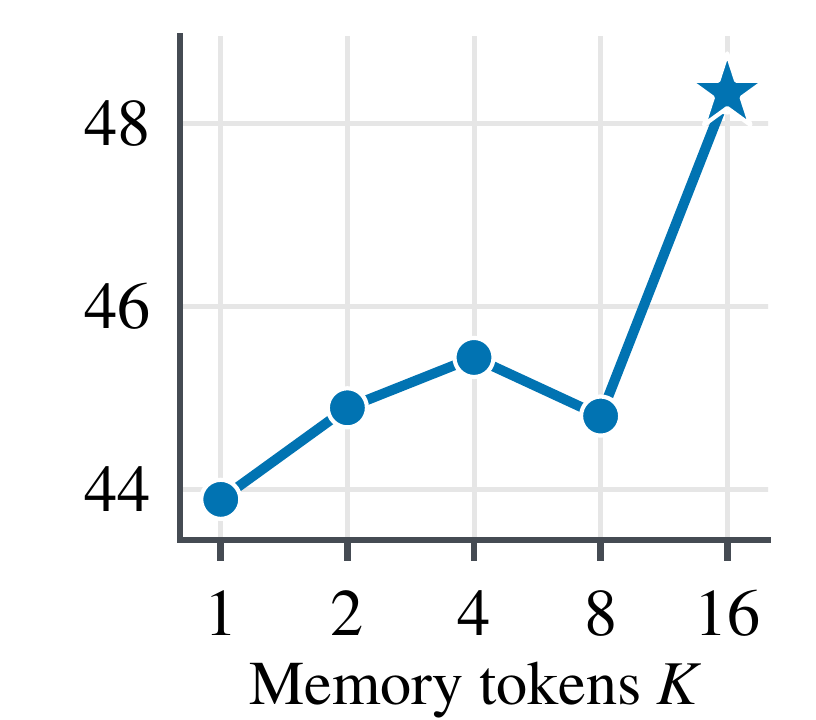}
  \caption{LaMP-7 (ROUGE-L)}\label{fig:token-budget-lamp7}
\end{subfigure}\par\medskip
\begin{subfigure}[t]{1.36in}\centering
  \includegraphics{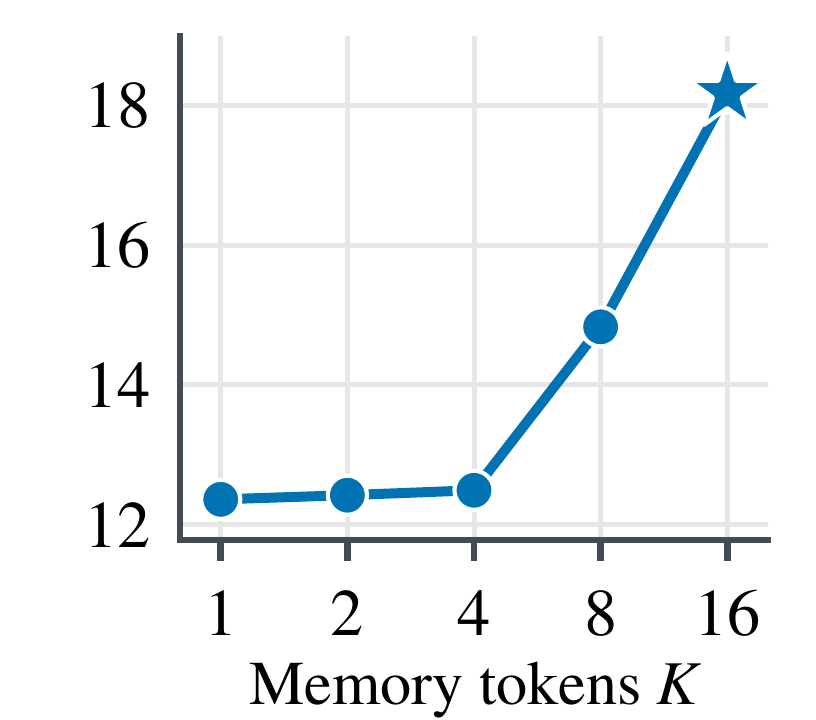}
  \caption{LaMP-4 (ROUGE-L)}\label{fig:token-budget-lamp4}
\end{subfigure}\hfill
\begin{subfigure}[t]{1.36in}\centering
  \includegraphics{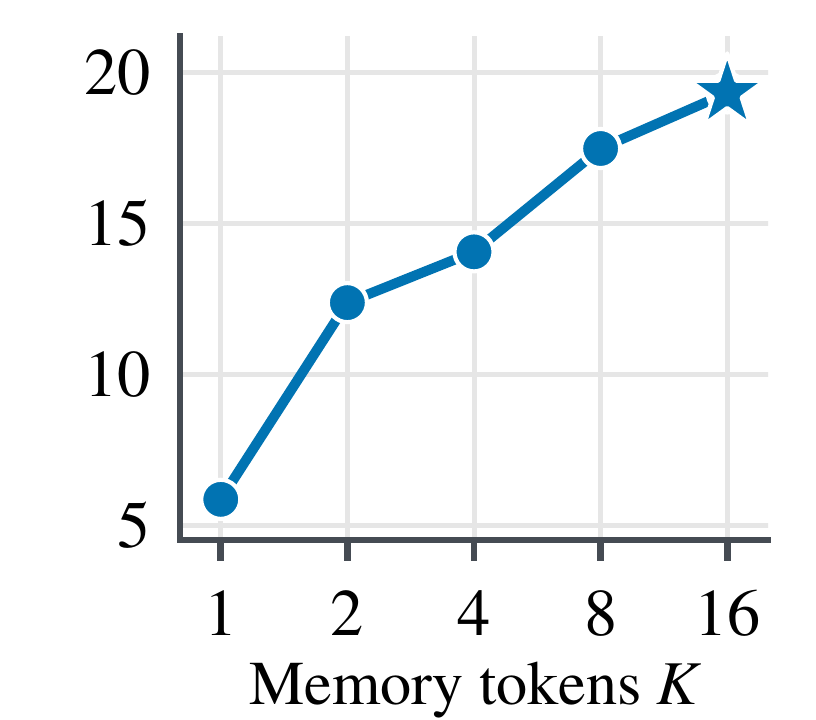}
  \caption{NQA (ROUGE-L)}\label{fig:token-budget-narrativeqa}
\end{subfigure}\hfill
\begin{subfigure}[t]{1.36in}\centering
  \includegraphics{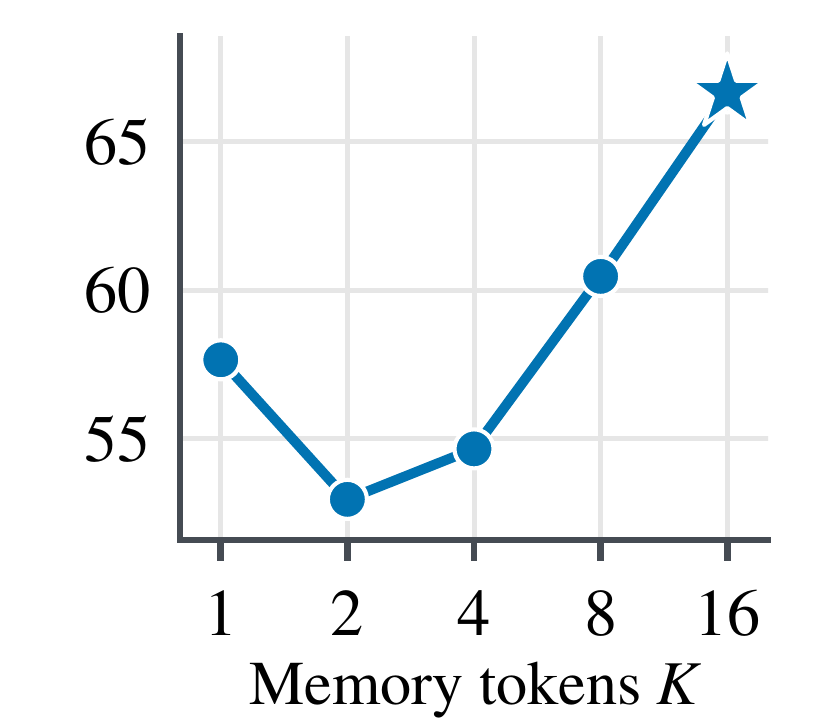}
  \caption{PV4 (Acc)}\label{fig:token-budget-pv4}
\end{subfigure}\hfill
\begin{subfigure}[t]{1.36in}\centering
  \includegraphics{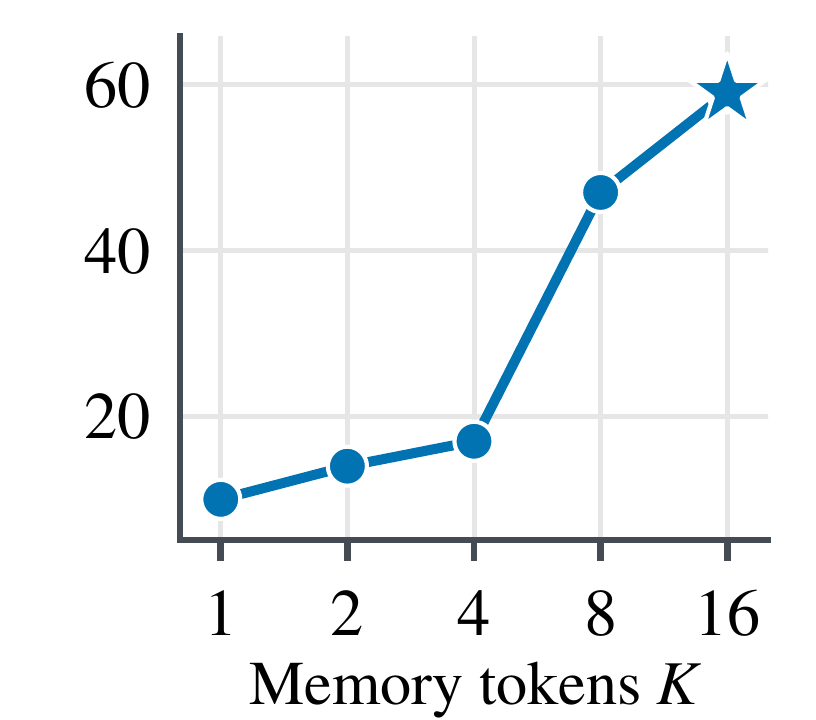}
  \caption{PubMedQA (Acc)}\label{fig:token-budget-pubmedqa}
\end{subfigure}
\caption{\textbf{Memory token budget at inference.} Performance of \memo with a frozen Qwen3-8B decoder using the first $K$ memory tokens of the checkpoints trained with the full $16$ tokens budget.
}
\label{fig:token-budget}
\end{figure}

Figure~\ref{fig:token-budget} varies the number of memory tokens $K$ at inference time without retraining. Both personalization benchmarks gain most from the last slots. PMv2 stays near its closed-book level through $K{=}4$ (26.58 against 25.64) and climbs to 44.60 at $K{=}16$, while PV4 starts near its closed-book level (57.66 against 58.30), dips at $K{=}2$ and $4$, and climbs to 66.65. Neither shows saturation, which reflects slots dropped from a checkpoint trained at $K{=}16$ rather than headroom above it, since training at $K{=}128$ does not help. FactKG, TriviaQA and NarrativeQA rise with every doubling. PubMedQA and LaMP-4 remain roughly flat at low $K$ and improve only from $K{=}8$, and LaMP-7 only from $K{=}16$. Across all three regimes, the required memory capacity tracks the degree of cross-document integration each task demands.

\subsection{Frozen vs. LoRA Training}

\begin{figure}[b!]
\centering
\includegraphics{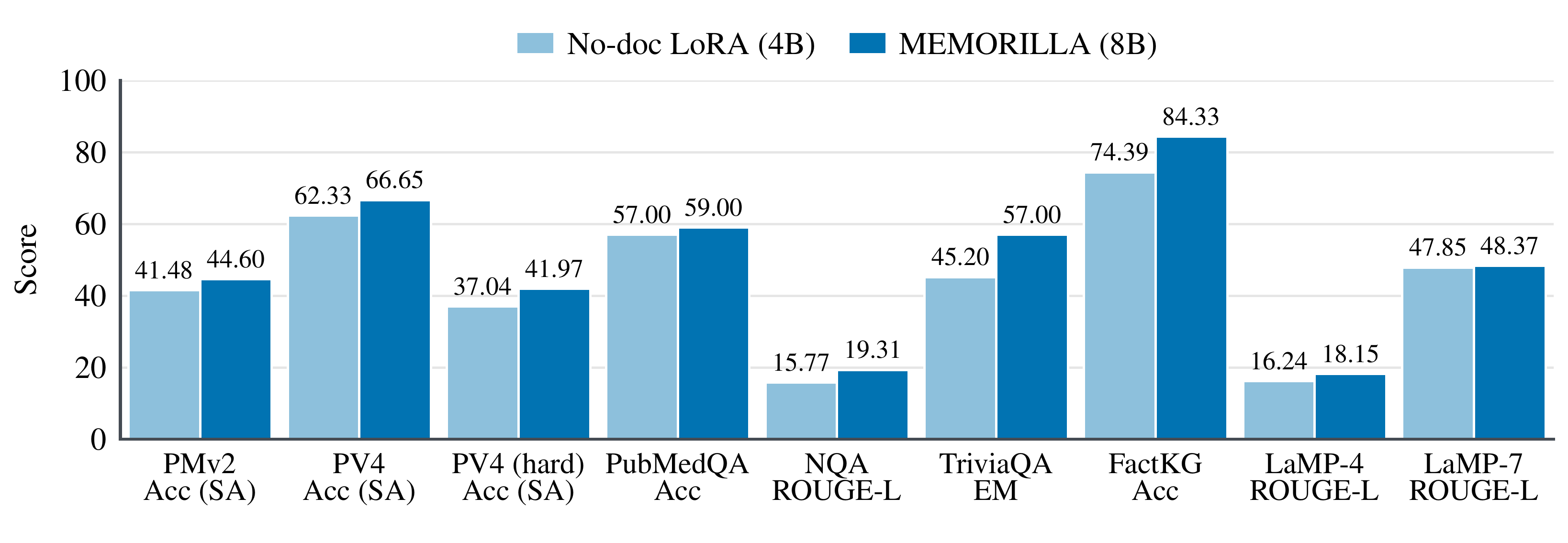}
\caption{\textbf{Memory versus comparable-budget adaptation.} \memo with an 8B decoder and 194.1M trainable parameters against a no-document rank-130 LoRA adapter with a 4B decoder and 268.4M trainable parameters.
}
\label{fig:lora}
\end{figure}

To separate the memory interface from generic adaptation, we train rank-130 LoRA adapters~\citep{lora} on the Qwen3-4B-Instruct-2507 decoder with 268.4M parameters, comparable to the 194.1M parameters of the memory module paired with the 8B decoder. The adapters use the same training rows as \memo's per-task fine-tuning and receive no documents in context. Figure~\ref{fig:lora} shows that \memo leads on every primary metric, although the PubMedQA and LaMP-7 gaps are within noise. Together with the noise test above, where \memo loses 5.93 on FactKG without real documents, this indicates that its gains rely on the stored evidence rather than on adaptation alone. Freezing both backbones keeps the store portable and the decoder untouched, at the cost of an encoder never optimized for compression.

\subsection{Role of Encoder Size}

\begin{figure}[h!]
\centering
\includegraphics{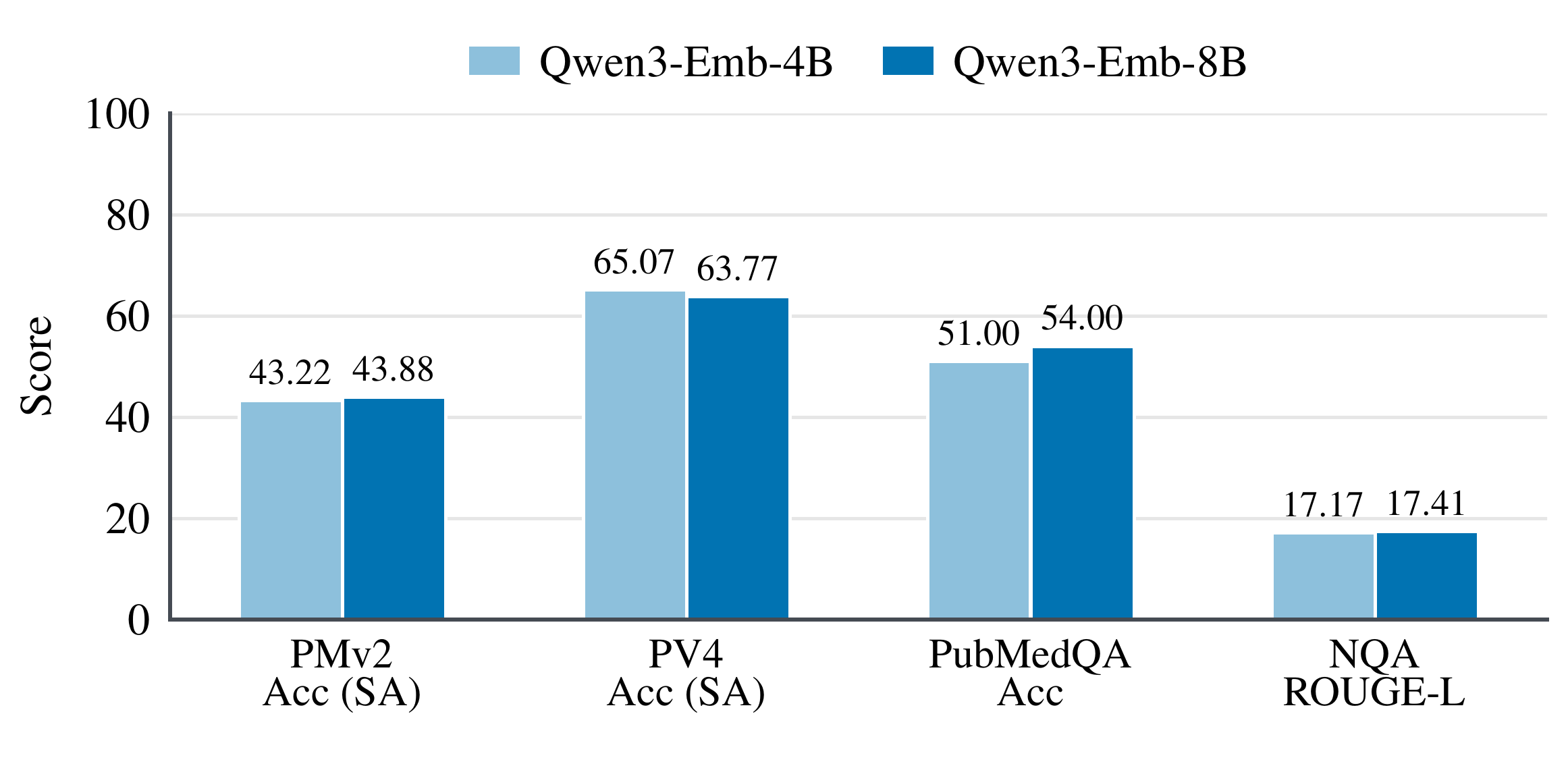}
\caption{\textbf{Encoder size.} The memory module is trained from scratch with a frozen Qwen3-8B decoder and either encoder; all other settings are fixed.}
\label{fig:encoder}
\end{figure}

Figure~\ref{fig:encoder} swaps the frozen document encoder from Qwen3-Embedding-4B to Qwen3-Embedding-8B, keeping the Qwen3-8B decoder, memory module and recipe fixed and training from scratch. Every score moves by at most 1.3 except PubMedQA, where the +3.00 gain corresponds to three of 100 questions, so a larger encoder from the same family has minimal effects on performance.

\section{Computational Analysis of \memo}
\label{sec:compt}

Consider a collection of $N$ documents serving $Q$ queries. \memo pays the encoder once on the write path, producing $N$ embeddings that persist and are reused by every later query. Dense retrieval builds the same index, so this cost is shared rather than added. The two methods differ on the read path. RAG places the selected text in the decoder, so its prefill grows with retrieval depth, while \memo runs one cross-attention read over the stored embeddings and always hands the decoder $K{=}16$ tokens.
Table~\ref{tab:compute} reports measurements on one H100-80GB at batch size 1 with a Qwen3-8B decoder. Growing the collection by 512$\times$ decreases the memory-module forward time by 0.06 ms and \memo end-to-end latency by 1.9 ms, while full-context decoding grows by 19{,}184 ms and exceeds the enforced 32{,}768-token limit, requiring YaRN. Transfer stays under 0.04\% of query latency because a 200 document collection is 1 MB, and the store is 7{,}266$\times$ smaller than the corresponding KV cache on PV4 (50{,}455 document tokens at 144 KiB per token for Qwen3-8B in bf16, 6.9 GiB, against 200 embeddings of 2{,}560 bf16 values). Absolute latency is dominated by decoding rather than by memory, so the benefit is not a faster single query but a serving cost that stays flat as collections grow and amortizes across the $Q$ queries that read them.

\begin{table}[h!]
\caption{\textbf{Serving cost as the collection grows.} One NVIDIA H100-80GB, batch size 1, bf16, greedy, \texttt{max\_new\_tokens}=32, 250-token documents; median of 5 timings after 2 warmups. The write path is paid once per document and is shared with dense retrieval; read path and decoder cost are paid per query. Full Context at $N{=}512$ is 128{,}064 context tokens, past the enforced 32{,}768-token limit, so it uses YaRN (factor 4). Timings use HF transformers \texttt{generate} without a paged KV cache, so Full Context latency depends on the serving stack.}
\label{tab:compute}
\centering
\scriptsize
\setlength{\tabcolsep}{0pt}
\begin{tabular*}{\textwidth}{@{\extracolsep{\fill}} l l r r r @{}}
    \toprule
    \textbf{Path} & \textbf{Quantity} & $N{=}1$ & $N{=}512$ & \textbf{Growth} \\
    \midrule
    \multirow[t]{2}{*}{\textbf{Write}}
        & Document encoding    & 22.92 ms & 2{,}588.84 ms & linear in $N$ \\
        & Storage per document & 5 KB     & 5 KB          & flat \\
    \cmidrule{1-5}
    \multirow[t]{2}{*}{\textbf{Read}}
        & Memory module forward & 2.01 ms  & 1.96 ms  & $-0.06$ ms \\
        & CPU to GPU transfer   & 0.014 ms & 0.060 ms & $+0.046$ ms \\
    \cmidrule{1-5}
    \multirow[t]{3}{*}{\textbf{Decoder}}
        & Context tokens            & 16     & 16            & flat \\
        & \memo{} end to end        & 740 ms & 739 ms        & $-1.9$ ms \\
        & Full Context end to end   & 749 ms & 19{,}932 ms   & $+19{,}184$ ms \\
    \bottomrule
\end{tabular*}
\end{table}

\section{Continual Learning with \memo: Additional Details}
\label{app:textworld-rl}

\begin{figure}[t!]
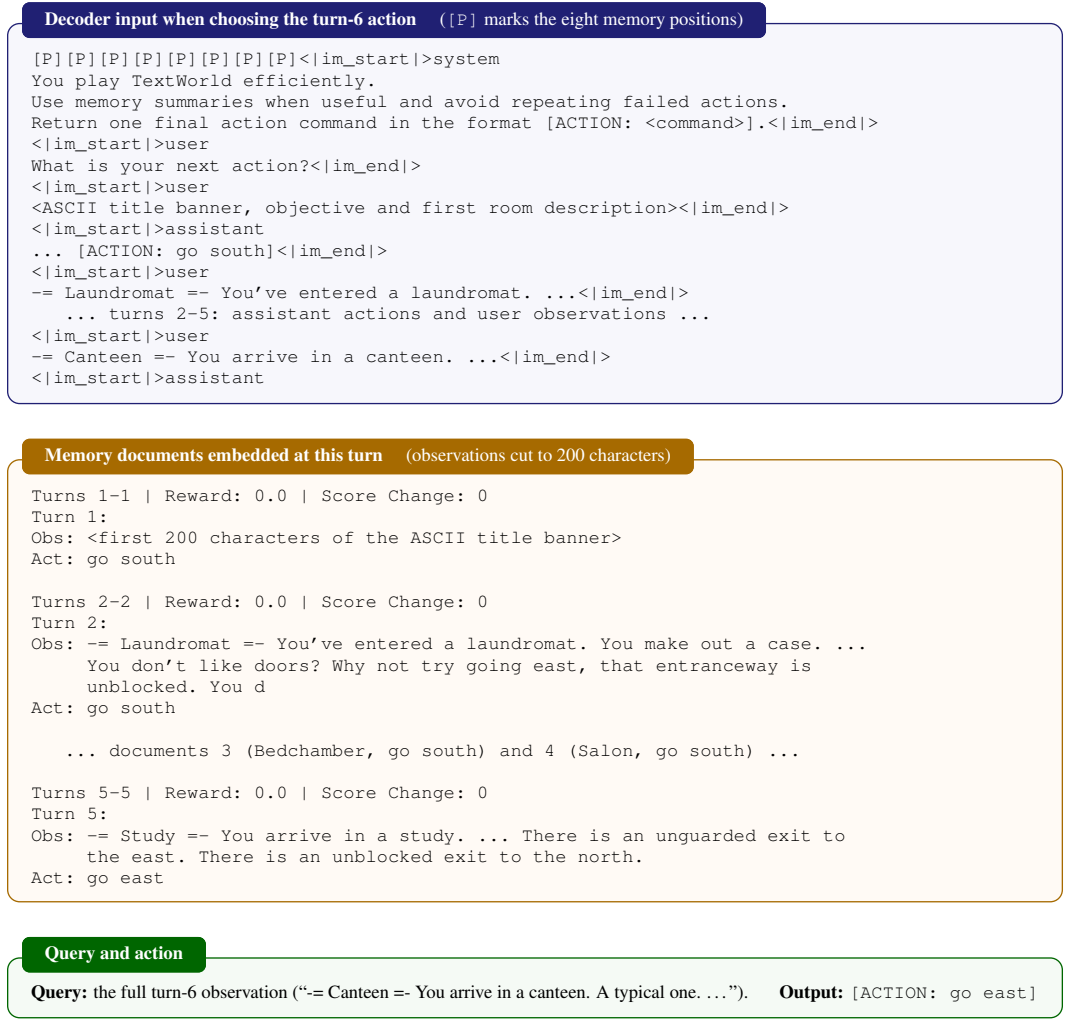

\centering
\begin{tcolorbox}[tw, colback=twblueback, colframe=twblueframe, colbacktitle=twblueframe,
  title={Decoder input when choosing the turn-6 action \quad \normalfont(\texttt{[P]} marks the eight memory positions)}]
\begin{verbatim}
[P][P][P][P][P][P][P][P]<|im_start|>system
You play TextWorld efficiently.
Use memory summaries when useful and avoid repeating failed actions.
Return one final action command in the format [ACTION: <command>].<|im_end|>
<|im_start|>user
What is your next action?<|im_end|>
<|im_start|>user
<ASCII title banner, objective and first room description><|im_end|>
<|im_start|>assistant
... [ACTION: go south]<|im_end|>
<|im_start|>user
-= Laundromat =- You've entered a laundromat. ...<|im_end|>
   ... turns 2-5: assistant actions and user observations ...
<|im_start|>user
-= Canteen =- You arrive in a canteen. ...<|im_end|>
<|im_start|>assistant
\end{verbatim}
\end{tcolorbox}

\vspace{2pt}
\begin{tcolorbox}[tw, colback=tworangeback, colframe=tworangeframe, colbacktitle=tworangeframe,
  title={Memory documents embedded at this turn \quad \normalfont(observations cut to 200 characters)}]
\begin{verbatim}
Turns 1-1 | Reward: 0.0 | Score Change: 0
Turn 1:
Obs: <first 200 characters of the ASCII title banner>
Act: go south

Turns 2-2 | Reward: 0.0 | Score Change: 0
Turn 2:
Obs: -= Laundromat =- You've entered a laundromat. You make out a case. ...
     You don't like doors? Why not try going east, that entranceway is
     unblocked. You d
Act: go south

   ... documents 3 (Bedchamber, go south) and 4 (Salon, go south) ...

Turns 5-5 | Reward: 0.0 | Score Change: 0
Turn 5:
Obs: -= Study =- You arrive in a study. ... There is an unguarded exit to
     the east. There is an unblocked exit to the north.
Act: go east
\end{verbatim}
\end{tcolorbox}

\vspace{2pt}
\begin{tcolorbox}[tw, colback=twgreenback, colframe=twgreenframe, colbacktitle=twgreenframe, title={Query and action}]
\textbf{Query:} the full turn-6 observation (``-= Canteen =- You arrive in a canteen. A typical one. \ldots''). \hfill
\textbf{Output:} \texttt{[ACTION: go east]}
\end{tcolorbox}
\caption{\textbf{An example TextWorld prompt.} A winning 12-turn rollout on a held-out medium game, at the choice of the turn-6 action; training prompts have the same format. The eight memory vectors sit before the system prompt and are recomputed every turn from one document per completed turn, with the current observation as the query, while the full trajectory stays in the prompt. Line breaks inside the documents are added for width.}
\label{fig:textworld-prompt}
\end{figure}

TextWorld games require an agent to maintain state across a partially observable trajectory, navigate a map, manipulate objects, and execute ordered action sequences. Observations are room descriptions and action outcomes, actions are short text commands, and games span up to 65 rooms. From the second turn on, \memo embeds each completed turn as one document, holding its observation truncated to 200 characters and its action (74 tokens on average), reads these documents with the current observation as the query, and places the resulting $K{=}8$ memory vectors before the prompt. The 20 extreme, huge, long and mega test games instead embed one document per five completed turns with full observations, keep the last four documents, and use the mean document embedding as the query. Figure~\ref{fig:textworld-prompt} shows a prompt and the documents compressed into the memory vectors, and Section~\ref{sec:textworld-setup} reports how the trajectory grows across a rollout.

We train on 800 prompts and evaluate on held-out validation and test splits of 40 games each (30 distinct, since the five medium and five hard games appear twice). Games come in six tiers (medium, hard, long, mega, huge, extreme) with 15--65 rooms and 6--74-step quests; the 800 training prompts are 202 medium, 558 hard and 10 from each longer tier. Rollouts sample up to 50 environment turns with at most 128 new tokens per turn, and evaluation samples three completions per held-out game at temperature 0.6 using the same SkyRL~\citep{cao2025skyrl} generation harness for every condition. The random-memory baseline uses the \memo architecture with untrained weights, so it adds eight latents to the same prompt without having learned what to place in them. Table~\ref{tab:textworld-config} lists the full configuration.

\subsection{Reinforcement-Learned Memory Objective}
\label{sec:rl_objective}

For agentic settings, we train the memory module from environment rewards while keeping the encoder and decoder frozen (Figure~\ref{fig:textworld_obj}). At each TextWorld turn, the interaction trajectory is embedded as a document collection, $f_{\text{mem}}$ maps it to memory vectors, and the frozen decoder samples the next action. We optimize only $\boldsymbol{\theta}_{\text{mem}}$ with Group Relative Policy Optimization (GRPO)~\citep{shao2024deepseekmath}, a critic-free variant of PPO~\citep{schulman2017ppo}. For each prompt, a group of $G$ rollouts receives scalar rewards $R_i$ from the environment, and advantages are standardized within the group:
\begin{equation}
    A_i = \frac{R_i - \operatorname{mean}_{j \in G}(R_j)}{\operatorname{std}_{j \in G}(R_j) + \epsilon}.
\end{equation}
The memory policy is then updated with the clipped PPO surrogate over generated action tokens,
\begin{equation}
    \mathcal{L}_{\text{RL}}(\boldsymbol{\theta}_{\text{mem}})
    = -\mathbb{E}\left[
    \min\left(r_i A_i,\; \operatorname{clip}(r_i, 1-\epsilon_c, 1+\epsilon_c) A_i\right)
    \right], \quad
    r_i = \frac{\pi_{\boldsymbol{\theta}_{\text{mem}}}(a_i \mid \tau_i)}{\pi_{\boldsymbol{\theta}_{\text{old}}}(a_i \mid \tau_i)}.
\end{equation}
The reward of each turn is the TextWorld game score after it, without shaping, and a rollout's return $R_i$ sums these over its turns. Most medium and hard games have a single one-point quest and end when it is solved, so they return 1 for a win and 0 otherwise; games in the four longer tiers (long, mega, huge and extreme) have one to three quests and also score as each is completed. The injection mechanism is unchanged from the supervised setting, while the memory module adds one self-attention layer over the trajectory documents before its cross-attention blocks. Memory here supplements the trajectory rather than replacing it, distinguishing it from the compression results in Section~\ref{sec:quantitative}. At turn $t$, the decoder input follows Eq.~(\ref{eq:memory_inj}) with the query replaced by the trajectory $\tau^{(t)}$, which contains the game description and all prior observations and actions:
\begin{equation}
    \mathbf{X}^{(t)}_{\text{LLM}}
    = [\mathbf{m}^{(t)}_1,\ldots,\mathbf{m}^{(t)}_K;\mathbf{P}_{\tau^{(t)}}],
\end{equation}
where the memory tokens $\mathbf{m}^{(t)}_k$ are recomputed at every turn after the first from the completed turns of $\tau^{(t)}$, with the current game observation as the query $q$ (the mean document embedding in the longer games).

\begin{table*}[h!]
\caption{\textbf{TextWorld RL configuration.} The memory module is the only trainable component, and the full textual history is retained in every condition for a fair comparison.}
\label{tab:textworld-config}
\centering
\scriptsize
\setlength{\tabcolsep}{3pt}
\renewcommand{\arraystretch}{1.10}
\begin{tabular*}{\textwidth}{@{\extracolsep{\fill}} l l l @{}}
    \toprule
    \textbf{Component} & \textbf{Setting} & \textbf{Value} \\
    \midrule
    \multirow[t]{4}{*}{\textbf{Environment}}
        & Task suite & TextWorld held-out games up to 65 rooms \\
        & Splits & \makecell[l]{800 train prompts; 40-prompt validation and test splits\\(30 games each)} \\
        & Agent state & Initial description plus prior observations/actions \\
        & Reward & TextWorld game score; no shaping reward \\
    \cmidrule{1-3}
    \multirow[t]{5}{*}{\textbf{Model interface}}
        & Decoder & Frozen Qwen3-4B-Instruct-2507 \\
        & Trajectory encoder & Frozen Qwen3-Embedding-4B \\
        & Memory budget & $K{=}8$ latent vectors inserted into decoder prompt \\
        & Memory architecture & 8 heads; 1 self-attention layer; 2 cross-attention layers \\
        & Context parity & \makecell[l]{Same input budget for every condition: 16{,}000 tokens in training,\\4{,}096 in evaluation; full history kept for ICL} \\
    \cmidrule{1-3}
    \multirow[t]{5}{*}{\textbf{Optimization}}
        & Trainable parameters & Memory module only; decoder and encoder frozen \\
        & Initialization & \makecell[l]{Random (Xavier-uniform slots); output projection $\mathcal{N}(0, 0.01)$;\\not loaded from the supervised checkpoint} \\
        & RL algorithm & GRPO; group size $G{=}16$; normalized group advantages \\
        & Policy objective & PPO clip 0.2; token-mean loss; no critic/KL/entropy \\
        & Optimizer & AdamW; learning rate $5{\times}10^{-4}$; no warmup \\
    \cmidrule{1-3}
    \multirow[t]{3}{*}{\textbf{Rollouts}}
        & Training batch & 4 prompts/step; 16 sampled rollouts/prompt \\
        & Horizon & Up to 50 environment turns/rollout \\
        & Action generation & Up to 128 new tokens/turn with SkyRL \\
    \cmidrule{1-3}
    \multirow[t]{2}{*}{\textbf{Evaluation}}
        & Sampling & 3 samples/held-out game at temperature 0.6 \\
        & Metrics & avg@3 mean solve rate; pass@3 any-sample solve rate \\
    \bottomrule
\end{tabular*}
\renewcommand{\arraystretch}{1.0}
\end{table*}

\end{document}